\documentclass[dvipsnames,sigconf=true, nonacm=true, review=false, anonymous = false]{acmart}
\AtBeginDocument{%
  }

\usepackage{pifont}
\usepackage{multirow}
\usepackage{adjustbox}
\usepackage{tcolorbox}
\usepackage{soul}               % For text highlighting and underlining
\usepackage{booktabs}           % For formal tables
\usepackage{colortbl}           % For colored table lines
\graphicspath{{pics/}}          % Path for graphics
\usepackage{tikz}
\usetikzlibrary{shapes.geometric}
\usetikzlibrary{positioning,fit,shapes,shadows,arrows,calc}
\usetikzlibrary{intersections}
\usepackage{etoolbox}           % Programming tools for conditional commands
\usepackage{amsmath}            % Math symbols and environments
\usepackage{amsfonts}           % Fonts like \mathbb for math
\usepackage{amsthm}             % Theorems and proofs
\usepackage{xspace}             % Smart spacing after commands
\usepackage{dsfont}             % Double-stroke fonts, e.g., \mathds{1}
\usepackage{rotating}           % Rotating text and figures
\usepackage{bm}                 % Bold math symbols
\usepackage{subcaption}         % Subfigure captions
\usepackage{hyperref}           % Hyperlinks for references
\usepackage{mathtools}          % Extensions for amsmath
\usepackage{url}                % Handling URLs properly
\usepackage[noabbrev,nameinlink]{cleveref} % Smart cross-references with links
\usepackage{bbm}

\usepackage{pgfplots}  % For creating plots
\usepackage{tikz}      % For general graphics, required for pgfplots
\usepackage{pgfplotstable}  % For reading and plotting from data files
\pgfplotsset{compat=1.15}
\pgfdeclarelayer{background}
\pgfdeclarelayer{foreground}
\pgfsetlayers{background,main,foreground}
\usepackage[skip=7pt]{caption}

\usepackage[algo2e,ruled,vlined,linesnumbered]{algorithm2e}

\usepackage{cleveref}

\definecolor{combinatorialcolor}{HTML}{708ec0}
\definecolor{factoredcolor}{HTML}{e49e75}

\tikzstyle{activity}=[align=center, rectangle, draw=black, rounded corners, text width=8em, fill=white, drop shadow]
\tikzstyle{data}=[align=center, rectangle, draw=black, fill=black!10, text width=8em, drop shadow]
\tikzstyle{internal}=[rectangle, draw=black!60, thick, align=left, text width=8em, minimum height=.75cm, text=black!60]
\tikzstyle{myarrow}=[-latex, thick]
\DeclareMathOperator*{\argmax}{arg\,max}

\newcommand{\R}{\mathbb{R}}

\definecolor{pastelred}{rgb}{1.0, 0.6, 0.6}
\definecolor{pastelgreen}{rgb}{0.6, 1.0, 0.6}
\definecolor{pastelblue}{rgb}{0.6, 0.6, 1.0}
\definecolor{pastelcyan}{rgb}{0.6, 1.0, 1.0}
\definecolor{pastelmagenta}{rgb}{1.0, 0.6, 1.0}
\definecolor{pastelyellow}{rgb}{1.0, 1.0, 0.6}
\definecolor{pastelorange}{rgb}{1.0, 0.8, 0.6}

\newcommand{\irace}{\textsc{irace}\xspace}
\newcommand{\onell}{(1+($\lambda$,$\lambda$))-GA\xspace}
\newcommand{\onemax}{\textsc{OneMax}\xspace}

\newcommand{\algocmt}[1]{\hfill\textcolor{lightgray}{$\triangleright$ #1}} % Define a custom comment command

\DeclareMathOperator{\flip}{flip}
\DeclareMathOperator{\cross}{cross}

\begin{document}

%\title{Multi-Parameter Control in Reinforcement Learning-Based Dynamic Algorithm Configuration: \\A Case Study on OneMax with (1+($\lambda$,$\lambda$))-GA}
\title[Multi-parameter Control for the $(1+(\lambda,\lambda))$-GA on OneMax via Deep Reinforcement Learning]{Multi-parameter Control for the $(1+(\lambda,\lambda))$-GA on OneMax \\via Deep Reinforcement Learning}

\author{Tai Nguyen}
\orcid{0009-0004-7707-2069}
\affiliation{
  \institution{University of St Andrews}
  \city{St Andrews}
  \country{United Kingdom}
}
\affiliation{
\institution{Sorbonne Universit\'e, CNRS, LIP6}
\city{Paris}
\country{France}
}

\author{Phong Le}
\orcid{0009-0000-0749-9519}
\affiliation{
  \institution{University of St Andrews}
  \city{St Andrews}
  \country{United Kingdom}}

\author{Carola Doerr}
\orcid{0000-0002-4981-3227}
\affiliation{
  \institution{Sorbonne Universit\'e, CNRS, LIP6}
  \city{Paris}
  \country{France}}

\author{Nguyen Dang}
\orcid{0000-0002-2693-6953}
\affiliation{
  \institution{University of St Andrews}
  \city{St Andrews}
  \country{United Kingdom}}

\renewcommand{\shortauthors}{Nguyen et al.}


\begin{abstract}

It is well known that evolutionary algorithms can benefit from dynamic choices of the key parameters that control their behavior, to adjust their search strategy to the different stages of the optimization process. A prominent example where dynamic parameter choices have shown a provable super-constant speed-up is the $(1+(\lambda,\lambda))$ Genetic Algorithm optimizing the OneMax function. While optimal parameter control policies result in linear expected running times, this is not possible with static parameter choices. This result has spurred a lot of interest in parameter control policies. However, many works, in particular theoretical running time analyses, focus on controlling one single parameter. Deriving policies for controlling multiple parameters remains very challenging. In this work, we reconsider the problem of the $(1+(\lambda,\lambda))$ Genetic Algorithm optimizing OneMax. We decouple its four main parameters and investigate how well state-of-the-art deep reinforcement learning techniques can approximate good control policies. We show that although making deep reinforcement learning learn effectively is a challenging task, once it works, it is very powerful and is able to find policies that outperform all previously known control policies on the same benchmark. Based on the results found through reinforcement learning, we derive a simple control policy that consistently outperforms the default theory-recommended setting by $27\%$ and the irace-tuned policy, the strongest existing control policy on this benchmark, by $13\%$, for all tested problem sizes up to $40{,}000$.
\end{abstract}

\begin{CCSXML}
<ccs2012>
<concept>
<concept_id>10010147.10010178.10010205.10010209</concept_id>
<concept_desc>Computing methodologies~Randomized search</concept_desc>
<concept_significance>500</concept_significance>
</concept>
</ccs2012>
\end{CCSXML}

\ccsdesc[500]{Computing methodologies~Randomized search}

\maketitle

\section{Introduction}

Parameter control, studying how to best adjust the parameters of an optimization algorithm during its execution, is a topic that has gained significant interest in the evolutionary computation community in the last decades, both from an empirical angle~\cite{KarafotiasSE12,EibenHM99,AletiM16} and from a running time analysis perspective~\cite{doerr2020theory}. Much of this interest was spurred by results showing that algorithms using a dynamic setup can significantly outperform their static counterparts. However, identifying optimal or reasonably good control policies remains a very challenging endeavor. 

For the theory community, an important example that spurred much of the recent interest in parameter control is the observation that the $(1+(\lambda,\lambda))$~Genetic Algorithm (GA)\footnote{We are aware that the reference to \emph{genetic algorithm} in the name may be ambiguous, but since the algorithm is well established, we prefer to refer to it by its original name.} equipped with a one-fifth-success rule has an expected linear running time on \onemax; a performance that is impossible to obtain with static parameter settings~\cite{doerr2018optimal}. Although such self-adjusting setting was proven beneficial, it remains limited to controlling a single parameter of the algorithm, the offspring population size $\lambda$. A decoupling of the key parameters of the \onell has been considered in~\cite{dang2019hyper} and, in a different way, in~\cite{AntipovBD24}. In the approach of~\citet{dang2019hyper}, the policy still controls one parameter, the offspring population size $\lambda_m$ used in the mutation phase, whereas the offspring size of the crossover phase $\lambda_c=\gamma \lambda_m$, the mutation rate $p=\alpha \lambda/n$, and the crossover bias $c=\beta/\lambda_c$ are parameterized in static dependence of this controlled parameter. The work~\citet{AntipovBD24} uses a different approach and samples, in each iteration, the values of the three parameters $\lambda_m$, $p$, and $c$ from heavy-tailed distributions. 

In this work, we take an alternative approach to discover multi-parameter control policies for the \onell on \onemax. Instead of regarding the traditional parameter control setting, where parameter values have to be chosen ``on the fly'' (or ``\emph{online}'', as one would refer to in a Machine Learning context), i.e., while optimizing the problem at the same time, we move our focus to the \emph{Dynamic Algorithm Configuration} setting~\cite{biedenkapp2022theory,adriaensen2022automated}, where state-dependent parameter settings are learned through a dedicated training phase. In this context, we investigate the potential of leveraging (deep) reinforcement learning (RL) to support finding effective multi-parameter control policies for the \onell on \onemax. In fact, RL fits very well with parameter control contexts where the aim is to learn a policy that maps from the current state of the algorithm (during the optimization process) to the best parameter values for that state. Deep RL has become increasingly popular in algorithmic communities for learning parameter control policies in evolutionary algorithms (e.g., ~\cite{sharma2019deep,tessari2022reinforcement,biedenkapp2022theory,ma2024auto}), for adaptively selecting components and adjusting them in hyper-heuristics (e.g.,~\cite{yi2022automated,yi2023automated}), and for dynamically selecting heuristics in AI planning solvers (e.g.,~\cite{speck2021learning}).

Despite their potentials, adopting deep-RL approaches in DAC context is non-trivial, as illustrated in recent studies on deep-RL learning stability using theory-derived DAC benchmarks~\cite{biedenkapp2022theory,nguyen2025importance}. Previous theory--derived deep-RL studies were limited to single--parameter control settings, while in this work, we address a multi-parameter control problem using the \onell on \onemax. Referring the interested reader to~\cite{chen2023using} for a detailed analysis of the problem landscape, we point out that even for this comparatively simple algorithm optimizing the artificial \onemax problem, the structure of the parameter control policy problem is non-trivial and imposes challenges when learning in the corresponding DAC landscape. We emphasize that our core interest is in the control problem, not in the optimization of the \onemax problem. Put differently, we use this benchmark because it is well explored and offers a well-motivated baseline against which we can compare.

Our main contributions can be summarized as follows:

\begin{enumerate}

    \item We leverage Double Deep Q-Network (DDQN)~\cite{mnih2013playing}, a commonly used deep RL approach in dynamic algorithm configuration contexts~\cite{sharma2019deep,speck2021learning,nguyen2025importance}, to control four key parameters of the \onell. As most, but not all~\cite{AntipovDK22,AntipovDK19,BuzdalovD17,GoldmanP15} previous works on the \onell, we consider the well-explored setting of minimizing the expected running time of the \onell on \onemax. We demonstrate the challenges of applying DDQN in this context and highlight three key designs to enable effective learning, including: (i) a well-chosen representation of the action space; (ii) reward shifting; and (iii) an increased discount factor. With all these components in place, our DDQN-based approach is able to learn policies that outperform all previously known control policies.
    
    \item Based on the results suggested by DDQN, we derive a fairly simple and interpretable yet extremely effective parameter control policy for \onell on \onemax. Our derived policy significantly outperforms both theoretical and empirical work on parameter control. The effectiveness and the simplicity of our derived policy opens up the potential for new theoretical findings on this benchmark.
    
\end{enumerate}

With these contributions, we expect to address researchers from at least two different domains. Theory-interested researchers may find our derived control policies and the ablation studies valuable. Empirically-oriented researchers with interest in RL may be interested in the challenges of adopting deep RL in algorithmic contexts and the insights provided by our findings on the impact of those key designs in DDQN's learning ability.

\textbf{Reproducibility:} Our code and data are available at~\cite{source}. 

\section{Background}

\subsection{The \texorpdfstring{\onemax}{OneMax} Problem}

We consider the problem of minimizing the average running time of the \onell on \onemax, a setting that has received considerable attention in the running time analysis community over the last years~\cite{doerr2015black,doerr2018optimal,AntipovBD22,AntipovBD24,BassinBS21}. 

Below, we refer to \onemax as the original \onemax function $\onemax: \{0,1\}^{n} \to \mathbb{R}, x \mapsto \sum_{i=1}^{n}x_i$, but, as well known, the performance of the \onell is strictly identical on all \onemax-like functions $\onemax_z: \{0,1\}^{n} \to \mathbb{R}, x \mapsto |\{1 \le i \le n | x_i=z_i\}$, $z\in\{0,1\}$, since they are isomorphic to the original $\onemax=\onemax_{(1,\ldots,1)}$ and the \onell is an unbiased algorithm in the sense of Lehre and Witt proposed in~\cite{LehreW12}. All results presented below therefore naturally apply to all generalized $\onemax_z$ functions, sometimes referred to as \emph{Mastermind} with two colors~\cite{doerr2014playing}.

\subsection{The \texorpdfstring{\onell}{(1+(L,L)) GA}}

Algorithm~\ref{alg:onell} presents a generalized version of the \onell, originally proposed in~\cite{doerr2015black}, with \emph{four parameters} to be controlled, namely the \emph{population sizes} $\lambda_m \in \mathbb{N}$ (for the mutation phase) and $\lambda_c\in \mathbb{N}$ (for the crossover phase), the \emph{mutation rate coefficient} $\alpha \in \mathbb{R}_{>0}$, and the \emph{crossover bias coefficient} $\beta \in \mathbb{R}_{>0}$. 
We briefly explain the different phases and parameters in the following paragraphs. 
 
Starting from an initial solution, the \onell runs iteratively until an optimal solution is found.\footnote{In practice, one would of course use or add a second stopping criterion; but, as common in running time analysis, we only consider time to optimum here in this work.} 
In each iteration, the algorithm has two main phases: the mutation phase and the crossover phase. 

\textbf{Mutation phase.} 
In the mutation phase, $\lambda_m$ offspring are generated by applying standard bit mutation to the current-best solution $x$. More precisely, a mutation strength $\ell$ is sampled from the binomial distribution $Bin_{>0}(n,p)$ that is conditioned on returning a value $\ell>0$ through rejection sampling; i.e., $\ell$ is sampled from $\text{Bin}(n,p)$ until a positive value is returned. Following the recommendations made in~\cite{doerr2015black,doerr2018optimal}, the \emph{mutation rate} $p$ is parametrized as $p=\alpha \lambda_m / n$. Each of the $\lambda_m$ offspring $x^{(i)}$ is then obtained by applying the $\flip_{\ell}$ operator to $x$, which flips exactly $\ell$ bits in $x$, chosen uniformly at random.

\textbf{Crossover phase.} 
The best among the $\lambda_m$ solutions obtained in the mutation phase, referred to as $x'$, is selected for the crossover phase; in case of ties a random one among the best offspring is chosen. It is worth noting that $x'$ may (and frequently will be) worse than the current incumbent solution $x$. The goal of the crossover phase is to \emph{repair} it, see~\cite{doerr2015black} for a detailed intuition behind the \onell. In a nutshell, the crossover acts as a \emph{genetic repair mechanism} that mitigates the potential loss of beneficial genes from the parent due to the \emph{aggressive} mutation rate while preserving them in the offspring. To this end, $\lambda_c$ offspring are generated by combining $x'$ with its parent $x$ through the biased crossover operator $\cross_c$, which creates a new search point by choosing, independently for each position $1 \le i \le n$, the entry of the second input with probability $c$ and choosing the entry from the first argument otherwise. We refer to $c$ as the \emph{crossover bias,} and we parametrize $c=\beta/\lambda_c$, again following the suggestions made in~\cite{doerr2015black,doerr2018optimal}. As an implementation detail, also in line with experiments made in previous work~\cite[Section~4.1]{AntipovBD22}, we only evaluate solutions $y^{(i)}$ that are different from both, $x$ and $x'$. This implies that the number of evaluations in this crossover phase can be smaller than $\lambda_c$. Finally, we note that the use of the crossover as repair mechanism is specific to this algorithm; crossover can have very different roles in evolutionary algorithms. 

\textbf{Selection.} The best of all offspring, $y$ (selected again uniformly at random among all best options in case of multiple best offspring), replaces the parent $x$ if it is at least as good as it. 

\begin{algorithm2e}[t]%
\caption{The $(1+(\lambda,\lambda))$-GA with four parameters to be controlled, including the population sizes $\lambda_m$ (for the mutation phase) and $\lambda_c$ (for the crossover phase), the mutation rate coefficient $\alpha$, and the crossover bias coefficient $\beta$. \\\textbf{Input:} Problem size $n$, fitness function $f(\cdot)$}
\label{alg:onell}
%\small
$x \gets$ a sample from $\{0,1\}^{n}$ chosen uniformly at random\;
\While{$x$ is not the optimal solution}{

Choose $\lambda_m$, $\lambda_c$, $\alpha$ and $\beta$ based on $f(x)$

% $\left< \lambda_{m}, \alpha, \lambda_c, \beta \right> =\pi(s)$\;
% $p = \alpha \lambda_m/n$; and $c = \beta/\lambda_c$\;
%%%%
\underline{\textbf{Mutation phase:}}\\
% \Indp
        $p = \alpha \lambda_m/n$; \hfill \algocmt{mutation rate} \\
	Sample $\ell$ from $\text{Bin}_{>0}(n,p)$\;
    % Set population size $\Lambda$ as $\round{\lambda}$\;
	\lFor{$i=1, \ldots, \lambda_m$}
         {$x^{(i)} \leftarrow \text{flip}_{\ell}(x)$; Evaluate $f(x^{(i)})$}
	Choose $x' \in \{x^{(1)}, \ldots, x^{(\lambda_m)}\}$ with $f(x')=\max\{f(x^{(1)}), \ldots, f(x^{(\lambda_m)})\}$ u.a.r.\;
% \Indm
\underline{\textbf{Crossover phase:}}\\
% \Indp
 $c = \beta/\lambda_c$; \hfill \algocmt{crossover bias} \\
 $\mathcal{Y} \gets \emptyset$\;
\For{$i=1, \ldots, \lambda_c$}
{$y^{(i)} \leftarrow \text{cross}_{c}(x,x')$\; 
\lIf{$y^{(i)} \notin \{x,x'\}$}{Evaluate $f(y^{(i)})$; Add $y^{(i)}$ to $\mathcal{Y}$}}
\underline{\textbf{Selection and update step:}}\\
%If exists, choose $y \in \{y^{(1)}, \ldots, y^{(\lambda)}\} \setminus \{x\}$ 
%with 
%$f(y)=\max\{f(y^{(1)}), \ldots, f(y^{(\lambda)})\}$ u.a.r.; otherwise $y \assign x$\;
Choose $y \in \{x'\}\cup\mathcal{Y}$ with 
    $f(y) = \max\{f(x'), \max_{y' \in \mathcal{Y}}f(y')\}$ u.a.r.\;
% \Indm
% \Indp
%\lIf{$f(y) > f(x')$}{$y \leftarrow y'$ \textbf{ else } $y \leftarrow x'$}
\lIf{$f(y)\ge f(x)$}{$x \leftarrow y$} 
% \Indm
}
\end{algorithm2e}

\subsection{Single-parameter Control Policies}
\label{sec:single-param}

In Algorithm~\ref{alg:onell}, we allow the parameters to depend on the quality (``\emph{fitness}'') of the current-best solution (cf. line~3). Our key interest is in identifying parameter settings that minimize the expected number of function evaluations that the \onell performs until it finds the optimal solution. In other terms, we seek \emph{policies} that assign to each fitness value $f(x) \in \{0,\ldots,n-1\}$ values for our four parameters $\lambda_m$, $\lambda_c$, $\alpha$, and $\beta$. 

In the early works for this problem, only $\lambda_m$ is controlled, and a static dependence of $\lambda_c =\lambda_m$ and static choice $\alpha = \beta = 1$ was used. A key result in~\cite[Theorem~8]{doerr2015black} states that, with these choices and setting $\lambda_m=\sqrt{n/(n-f(x))}$, the \onell achieves a linear expected running time; a performance that is impossible to obtain with any fixed choices of $\lambda_m$, $\lambda_c$, $\alpha$, and $\beta$~\cite[Section~5]{doerr2018optimal}. 
% It was shown in~\citet{doerr2015black} that a well-configured $\lambda$ based on the fitness function resulted in better performance on the expected runtime compared to the optimal static setting. 

% Inspired by the idea of a single parameter configuration and Theorem 8 in \citet{doerr2015black}, Theorem 5 in \citet{doerr2020theory} states that the expected optimization time of \onell, with $p=\frac{\lambda}{n}$, $c=\frac{1}{\lambda}$, and $\lambda=\sqrt{n/(n-f(x))}$ on \onemax is $\Theta (n)$. 
Since such a precise dependence of $\lambda_m$ and $f(x)$ is difficult to guess without considerable effort, already in~\cite{doerr2015black}, a self-adjusting choice of $\lambda_m$ was experimented with. More precisely, the empirical results from~\cite{doerr2015black} suggest that a variant of the so-called \emph{one-fifth success rule}, originally proposed in the context of evolution strategies for continuous optimization~\cite{Devroye72,Rechenberg,SchumerS68} and adapted to discrete optimization settings in~\cite{KernMHBOK04}, leads to surprisingly good performance. That it indeed leads to an expected linear running time was then formally proven in~\cite[Theorem~9]{doerr2018optimal}. 
For later reference in this work, we provide the details of this control policy in Algorithm~\ref{alg:static_tune_onell}.

The self-adjusting \onell differs from Algorithm~\ref{alg:onell} in that we update the offspring population size $\lambda_m$ at the end of each iteration. If a better solution is found, the population size is increased to $A \lambda_m$, for some static update strength $A >1$. The population size is updated to $\beta \lambda_m$ otherwise, with $0<\beta<1$ being the second update strengths. To obtain a one-fifth-success rule, one sets $A=F^{(1/4)}$ and $b=1/F$ for some fixed factor $F>1$. With this setting, $\lambda_m$ remains constant if on average one out of five iterations is successful in finding a strictly better solution. This is the version investigated in~\cite{doerr2015black,doerr2015optimal}, with a fixed setting of $\alpha=\beta=\gamma=1$ in the notation of Algorithm~\ref{alg:static_tune_onell}.

In an attempt to automatize the identification of good parameter control policies, \citet{chen2023using} considered Algorithm~\ref{alg:onell} with $\lambda_c=\lambda_m$ and $\alpha=\beta=1$ to investigate how well the automated algorithm configuration tool \texttt{irace}~\cite{lopez2016irace} would approximate a good policy for controlling $\lambda_m$. While a na\"ive application of \texttt{irace} was unable to find good policies, an iterative ``binning-and-cascading'' approach was able to learn control policies that are significantly better than all previously known policies. However, the iterative configuration approach is computationally expensive, leading~\citet{nguyen2025importance} to investigate the effectiveness of employing deep reinforcement learning (RL), particularly DDQN~\cite{van2016deep}, to dynamically control $\lambda_m$, resulting not only in highly effective control policies but also a significant improvement in the sampling efficiency of the learning step compared to~\cite{chen2023using}.

\subsection{Multi-parameter Control Policies}
\label{sec:multi-param}

As already pointed out, Algorithm~\ref{alg:static_tune_onell} has more parameters than the algorithm originally proposed in~\cite{doerr2015black,doerr2018optimal}. This generalization was investigated in~\cite{dang2019hyper}, with the goal to understand the potential of fine-tuning the dependencies between the different parameters. \citet{dang2019hyper} tune all five parameters $\alpha$, $\gamma$, $\beta$, $A$, and $b$ using \texttt{irace}. Note that $\alpha$ and $\beta$ play the same roles as in Algorithm~\ref{alg:onell}, while parameter $\gamma$ introduces flexibility (compared to the original \onell) in choosing $\lambda_c=\gamma \lambda_m$ for $\gamma \in \R$. 

\begin{algorithm2e}[t]%
\caption{The self-adjusting $(1+(\lambda,\lambda))$-GA variant with five parameters $(\alpha, \gamma, \beta, A, b)$ to be tuned \emph{statically}~\cite{dang2019hyper}. The parts that are different from Algorithm~\ref{alg:onell} are marked in \textcolor{blue}{blue}. \\
\textbf{Input:} Problem size $n$, fitness function $f(\cdot)$, \textcolor{blue}{parameter values for $(\alpha, \gamma, \beta, A, b)$}
}
\label{alg:static_tune_onell}
%\small
$x \gets$ a sample from $\{0,1\}^{n}$ chosen uniformly at random\;
\textcolor{blue}{$\lambda_m \leftarrow 1$;} \\
\While{$x$ is not the optimal solution}{

\underline{\textbf{Mutation phase:}}\\
% \Indp
% \textcolor{blue}{$\lambda_m = \lambda$;} \hfill \algocmt{mutation population} \\
$p = \alpha \lambda_m / n$; \hfill \algocmt{mutation rate} \\
	Sample $\ell$ from $\text{Bin}_{>0}(n,p)$\;
    % Set population size $\Lambda$ as $\round{\lambda}$\;
	\lFor{$i=1, \ldots, \lambda_m$}
         {$x^{(i)} \leftarrow \text{flip}_{\ell}(x)$; Evaluate $f(x^{(i)})$}
	Choose $x' \in \{x^{(1)}, \ldots, x^{(\lambda_m)}\}$ with $f(x')=\max\{f(x^{(1)}), \ldots, f(x^{(\lambda_m)})\}$ u.a.r.\;
% \Indm
\underline{\textbf{Crossover phase:}}\\
% \Indp
\textcolor{blue}{$\lambda_c = \gamma \lambda_m$;} \hfill \algocmt{crossover population size} \\
$c = \beta/\lambda_c$; \hfill \algocmt{crossover bias} \\
$\mathcal{Y} \gets \emptyset$\;
\For{$i=1, \ldots, \lambda_c$}
{$y^{(i)} \leftarrow \text{cross}_{c}(x,x')$\; 
\lIf{$y^{(i)} \notin \{x,x'\}$}{Evaluate $f(y^{(i)})$; Add $y^{(i)}$ to $\mathcal{Y}$}
% \lIf{$y^{(i)} \notin \{x,x'\}$}{evaluate $f(y^{(i)})$}
}
%If exists, choose $y \in \{y^{(1)}, \ldots, y^{(\lambda)}\} \setminus \{x\}$ 
%with 
%$f(y)=\max\{f(y^{(1)}), \ldots, f(y^{(\lambda)})\}$ u.a.r.; otherwise $y \assign x$\;
\underline{\textbf{Selection and update step:}}\\
Choose $y \in \{x'\}\cup\mathcal{Y}$ with 
    $f(y) = \max\{f(x'), \max_{y' \in \mathcal{Y}}f(y')\}$ u.a.r.\;
% \Indm
% \Indp
\If{$f(y) > f(x)$}{
    $x \leftarrow y$; \\
    \textcolor{blue}{$\lambda_m \leftarrow \text{max}(A\lambda_m,1)$;} \hfill \algocmt{increasing $\lambda_m$} \\
}
\If{$f(y) = f(x)$}{
    $x \leftarrow y$; 
    \textcolor{blue}{$\lambda_m \leftarrow \text{max}(b\lambda_m,n-1)$;} \hfill \algocmt{decreasing $\lambda_m$} \\
}
\If{$f(y) < f(x)$}{
    \textcolor{blue}{$\lambda_m \leftarrow \text{min}(b\lambda_m,n-1)$;} \hfill \algocmt{decreasing $\lambda_m$} 
} 
}
\end{algorithm2e}

Based on results suggested by \texttt{irace},~\citet{dang2019hyper} were able to derive a multi-parameter control policy that outperforms existing single-parameter control policies for the same benchmark. Their policy suggested $\alpha=0.3594$, $\beta=1.4128$, $\gamma=1.2379$, $A=1.1672$, $b=0.691$. Their results confirm the significant performance gain when all four parameters of \onell are allowed to be chosen independently; some empirical results can be found in the subsequent sections. However, there are a number of limitations in their work. First, the derived policy required the algorithm to be slightly modified: the choices of $\lambda_m$ and $\lambda_c$ at each iteration are based on the fitness values of both the current solution and the solution obtained from the previous iteration, which may complicate theoretical analysis. Second, $\alpha$ and $\beta$ were selected statically, leaving an open question of whether there is potential gain when they are controlled dynamically. %Third, their method is quite computationally expensive, where each tuning consumes 240 CPU hours. 

In this work, we will take an alternative approach where we formulate the multi-parameter control task as Dynamic Algorithm Configuration (DAC)~\cite{biedenkapp2020dynamic}, where the aim is to leverage deep-RL to learn a policy that dynamically controls all four parameters $\lambda_m$, $\lambda_c$, $\alpha$, and $\beta$ without having to turn it into a static algorithm configuration problem. This would allow maximum flexibility in controlling those parameters. As we show in the later sections, our approach allows the derivation of a simple, yet highly effective policy that significantly outperforms any existing policies on this benchmark, including the one derived in~\citet{dang2019hyper}. 

\section{Deep RL for Multi-parameter Control of \onell on \onemax}

Our objective is to dynamically control four parameters $\lambda_{m}$, $\alpha$, $\lambda_c$, $\beta$ as illustrated in Algorithm~\ref{alg:onell}. Inspired by the work of~\citet{nguyen2025importance}, which demonstrates the potential of RL for single-parameter configuration on the same benchmark, we extend this line of research by formulating the multi-parameter configuration problem as an RL task ~\cite{sharma2019deep,speck2021learning} as below.

During the execution of the algorithm, a state $s \in \mathcal{S}$ captures all relevant information about its current status. An action $a \in \mathcal{A}$ specifies a tuple of values for the parameters to be used in the next iteration--namely, $\left< \lambda_{m}, \alpha, \lambda_c, \beta \right>$. Upon selecting an action $a$ (i.e., setting parameter values), the algorithm receives a reward $r = \mathcal{R}(s, a)$ that quantifies the quality of the chosen action in the given state.

This setting naturally fits into an RL framework, where the objective is to learn a policy $\pi^*$ that maps each state $s$ to an action $a = \pi^*(s)$ in order to maximize the expected cumulative reward over time:
\begin{equation}
\pi^* = \arg\max_{\pi} \mathbb{E}_\pi \left[ \sum_{k=0}^{\infty} \gamma^k \mathcal{R}(s_{t+k}, a_{t+k}) \,\middle|\, s_t = s, a_t = a \right],
\end{equation}
where $t$ is the current time step and $\gamma \in [0, 1)$ is the discount factor that balances immediate versus future rewards.

To deal with a multi-dimensional action space, a natural choice for the RL algorithm is Proximal Policy Optimization (PPO)~\cite{schulman2017proximal}, a versatile and well-known deep RL algorithm. Unfortunately, our preliminary experiments with PPO did not yield competitive results, even after extensive hyper-parameter optimization has been conducted. On the other hand, we were encouraged by the strong performance of the Dual Deep $\mathcal{Q}$-Network (DDQN) framework~\cite{mnih2013playing} reported in~\citet{nguyen2025importance} for single-parameter control on the same benchmark. Motivated by these findings, we adopt DDQN to address the more challenging multi-parameter configuration task. Briefly, DDQN is an RL algorithm that uses a neural network to approximate the action-value function $\mathcal{Q}(s, a)$, which estimates the expected return of taking action $a$ in state $s$ and following the policy thereafter.
\begin{figure*}[t]
    \centering
        \includegraphics[width=0.85\linewidth, trim=0 10pt 0 10pt, clip]{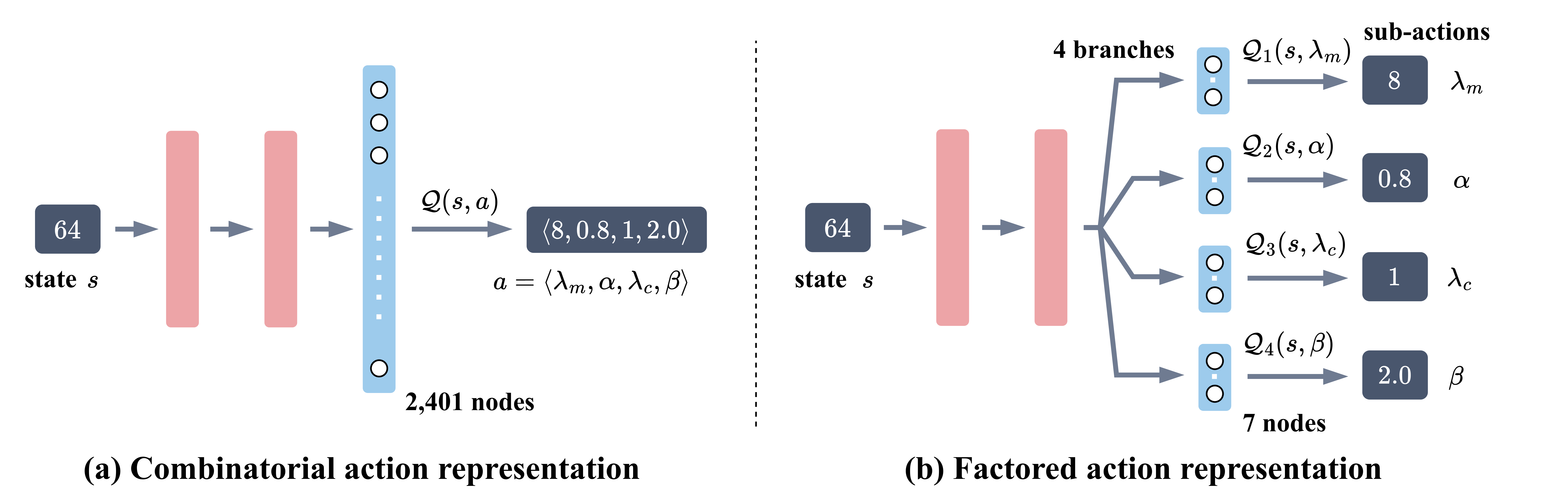}
    
    \caption{Deep $\mathcal{Q}$-network architectures for (a) combinatorial and (b) factored action space representation for \onell with four controllable parameters, each of which is selected from a set of 7 possible values. The number of output nodes for the combinatorial representation is $7^4=2{,}401$.}
    \label{fig:mp_dac}
\end{figure*}

\subsection{DDQN with Combinatorial Actions}
\label{sec:DQN_with_combinatorial_actions}
When the four parameters $\left< \lambda_{m}, \alpha, \lambda_c, \beta \right>$ are treated as a single atomic action, standard DDQN (as well as other RL methods) can be directly applied without requiring special treatment for the internal structure of the action space. In this work, we build on the approach proposed in~\citet{nguyen2025importance}, with extensions to support joint multi-parameter configuration.

First, to assess the difficulty of exploration in this combinatorial action space, we explore multiple reward functions proposed in \citet{nguyen2025importance}, beginning with the conventional reward function: $r_t = -E_t + \Delta f_t$. We then apply their adaptive shifted variant $r_t = -E_t + \Delta f_t + b^{-}_{a}$, where $E_t$ is the total number of solution evaluations in the current iteration, $\Delta f_t = f(x_{t+1}) - f(x_t)$ measures the fitness improvement, and $b^{-}_{a}$ is an adaptive bias term computed from the replay buffer during the warm-up phase to stabilize learning.

Second, we investigate the influence of the discount factor $\gamma$ on learning performance. As established in prior work~\cite{watkins1992q,712192,hu2022role}, the discount factor plays a critical role in reward estimation, which has been shown to be particularly sensitive in the \onemax benchmark~\cite{nguyen2025importance}. A larger $\gamma$ places greater emphasis on long-term rewards, potentially encouraging more strategic behavior but at the cost of slower convergence and increased instability. Conversely, a smaller $\gamma$ prioritizes immediate rewards, facilitating faster learning but risking shortsighted policies—especially in problems with delayed or sparse rewards.

The \onemax problem can be modeled as an \emph{episodic task}, where the optimization process is naturally segmented into episodes, each starting from a defined initial state and terminating either upon reaching the global optimum or hitting a predefined time cutoff. As recommended in~\citet{712192}, high or even undiscounted reward estimations are preferable in episodic tasks to avoid underestimating future gains. Accordingly, we evaluate two values for the discount factor: $\gamma \in \{0.99, 0.9998\}$, with the latter providing minimal discounting to better capture long-horizon reward signals.

Third, we modify the $\mathcal{Q}$-network architecture proposed in~\citet{nguyen2025importance} (illustrated in Figure~\ref{fig:mp_dac}a) such that its output layer spans the full combinatorial action space. At inference time, given a state $s$, the agent selects the optimal action according to:
\begin{equation}
a = \argmax_{\left< \lambda_{m}, \alpha, \lambda_c, \beta \right>} \mathcal{Q}(s,\left< \lambda_{m}, \alpha, \lambda_c, \beta \right>)
\end{equation}

\subsection{DDQN with Factored Actions}
\label{sec:DQN_with_factored_actions}

Treating each unique combination of sub-actions (i.e., individual parameter values) as a distinct atomic action—without considering the internal structure—is inefficient, especially when each sub-action can take on a large number of possible values. Specifically, if there are $n_d$ possible values for each sub-action and an action is composed of $D$ sub-actions $\langle a_1, \ldots, a_D \rangle$, then the total action space grows exponentially as $\prod_{d=1}^{D} n_d$. Such a combinatorial explosion poses serious challenges for RL, particularly for value-based methods like DDQN, as it hinders learning efficiency and increases inference cost~\cite{dulac2019challenges, covington2016deep, dulac1512deep, ddpg-soft-update}. To address this issue, prior work has explored hierarchical action selection~\cite{he-etal-2016-deep-reinforcement}, common-sense constraints~\cite{rasheed2020deep}, and learning-based reduction strategies~\cite{zahavy2018learn} to effectively prune the action space.

In our second RL approach, we leverage the structure of factored actions by decomposing each action into $D$ sub-actions, following methods from e.g., ~\citet{tavakoli2018action}. Each sub-action is modeled independently using a dedicated architecture. Concretely, for each dimension $d$, we define a distinct action-value function $\mathcal{Q}_d(s, a_d)$ that estimates the expected return of taking sub-action $a_d$ in state $s$, according to a local policy $\pi_d$. These functions are trained independently and capture only the marginal contribution of each dimension, without explicitly modeling interactions or dependencies between sub-actions.

This decomposition dramatically reduces the effective action space size and enhances scalability. To operationalize this structure, we adopt the \emph{action branching} framework introduced by~\citet{tavakoli2018action}, which offers an efficient and empirically validated approach for learning in factored action spaces. In our implementation, we construct a shared neural network backbone (illustrated in Figure~\ref{fig:mp_dac}b) with $D = 4$ output branches (or ``heads''), each corresponding to a sub-action-specific $\mathcal{Q}_d$ function. At inference time, the final joint action $a$ is constructed by independently selecting the optimal value in each sub-action dimension:
\begin{equation}
a=\left\langle
\begin{array}{l}
    \argmax_{\lambda_{m}}\mathcal{Q}_1(s,\lambda_{m})\\
    \argmax_{\alpha}\mathcal{Q}_2(s,\alpha)\\
    \argmax_{\lambda_{c}}\mathcal{Q}_3(s,\lambda_{c})\\
    \argmax_{\beta}\mathcal{Q}_4(s,\beta)\\
\end{array}
\right\rangle
\end{equation}

\subsection{Experimental Setup}
\label{sec:experiment_setup}

Because DDQN requires action spaces to be discrete, we first need to discretize the domains of \onell's parameters. Following ~\citet{nguyen2025importance}, we define the domain of $\lambda_m$ and $\lambda_c$ as $\mathcal{D}_1 = \{1, 2, 4, 8, 16, 32, 64 \}$. The upper bound was chosen based on results suggested in~\citet{nguyen2025importance}, where it was observed that the population size in the best policies learned by deep-RL never exceeded 64 for all tested problem sizes of $n \le 2000$. For $\alpha$ and $\beta$, we use $[0.25,2]$ as the domain following the setting in~\citet{dang2019hyper}. We discretize this domain uniformly for our DDQN experiments. For simplicity, we use the same number of values as in $\mathcal{D}_1$, i.e., $\alpha$ and $\beta$ are chosen from the set of $\mathcal{D}_2 = \{0.25, 0.542, 0.833, 1.125, 1.417, 1.708, 2\}$.

Each deep $\mathcal{Q}$-network has 2 hidden layers, each with 50 nodes. For the combinatorial representation, the output layer is comprised of $7^4=2{,}401$ output nodes. 

In line with \citet{biedenkapp2022theory,nguyen2025importance}, we use a standard DDQN hyperparameter setting with $\epsilon$-greedy exploration strategy where $\epsilon=0.2$. The replay buffer is configured to store 1 million transitions. Prior to training, we initialize the replay buffer by randomly sampling $10{,}000$ transitions. For optimization, we employ the Adam optimizer~\cite{adam} with a (mini-)batch size of $2{,}048$ and a learning rate of $0.001$. A cutoff time of $0.8 n^2$ solution evaluations, where $n$ is the problem size, is imposed on each \onell run to avoid wasting training budget on bad policies. 
%To align the two $\mathcal{Q}$-networks in DDQN (i.e., the target and the online networks), we apply the soft target update strategy~\cite{ddpg-soft-update} with an update factor of $0.01$ to coordinate the online and target policies \cite{ddpg-soft-update}. 

RL training is conducted per problem size with a training budget of $200{,}000$ training time steps for each RL training run. Following previous work~\citet{nguyen2025importance} on the same benchmark, to select the best policy from each RL training, we evaluate the learned policy at every $2{,}000$ training time steps with a budget of $100$ runs per evaluation. At the end of the training, we collect the top 5 best learned policies, re-evaluate each of them $1{,}000$ times and extract the best one. As deep-RL is commonly known to be unstable~\cite{henderson2018deep,islam2017reproducibility}, and since our main focus in this work is to leverage deep-RL to discover new parameter control policies for this particular benchmark, for each RL setting, we repeat the training $5$ times and use the best policy among those trainings as the final policy. 

All experiments are conducted on a machine equipped with a single-socket AMD EPYC 7443 24-Core Processor. For the RL training phase we use a single core, while for the evaluation phase, we parallelize 16 threads. The total wall-time for each RL training ranges between 30 minutes to one hour, depending on the problem size. This budget is smaller than the \texttt{irace}-based approach where a tuning was reported to take up to 2 days with 20 parallel threads~\cite{dang2019hyper}.

\textbf{Performance metric.} In the rest of this paper, the performance of each parameter control policy is reported using the Expected Runtime (ERT) across $1000$ random seeds. The runtime here is defined as the total number of solution evaluations used during a run of the \onell.

\textbf{Statistical tests.} In all result tables, the policy with the best ERT is highlighted in \textbf{bold}. We perform a Wilcoxon signed-rank test (confidence level $0.99$) comparing each of the remaining policies against the best one. Policies that are not statistically significantly different from the best are \underline{underlined}. We also apply the Holm–Bonferroni method to correct for multiple comparisons.

\subsection{Results}
\label{sec:ddqn_for_onemaxmpdac}

\begin{table*}[ht]
    \centering
\caption{ERT (and its standard deviation) normalized by the problem size $n$ across 1000 seeds of the best policies achieved by each combination of  \textcolor{combinatorialcolor}{\textbf{C}}ombinatorial and \textcolor{factoredcolor}{\textbf{F}}actored action representations (AR), Na\"ive and Adaptive Shifting (AS) reward functions, and discount factors ($\gamma$), across three problem sizes $n \in \{100, 200, 500\}$.}
    \label{tab:compare_rl_reward_gamma}
    %\begin{adjustbox}{max width=0.9\textwidth} % Scale the box to fit the page width
    \begin{tabular}{cccccc}
        \toprule
         % \multicolumn{3}{c}{\textbf{Setting}} & \multicolumn{3}{c}{\textbf{Problem Size}} \\
        % \midrule
        \textbf{AR} & \textbf{Reward} & $\mathbf{\gamma}$ & $\mathbf{n=100}$ & $\mathbf{n=200}$ & $\mathbf{n=500}$ \\
        \midrule 
\textcolor{combinatorialcolor}{\textbf{C}}&Na\"ive&$0.99$&4.510\scriptsize{(1.07)}&5.057\scriptsize{(0.83)}&5.959\scriptsize{(0.79)}\\
\textcolor{combinatorialcolor}{\textbf{C}}&AS&$0.99$&4.392\scriptsize{(0.77)}&4.961\scriptsize{(0.76)}&5.583\scriptsize{(0.74)}\\
\textcolor{combinatorialcolor}{\textbf{C}}&Na\"ive&$0.9998$&4.350\scriptsize{(0.80)}&4.969\scriptsize{(0.72)}&6.531\scriptsize{(0.61)} \\
\textcolor{combinatorialcolor}{\textbf{C}}&AS&$0.9998$&4.777\scriptsize{(0.81)}&4.835\scriptsize{(0.59)}&5.198\scriptsize{(0.51)}\\
\textcolor{factoredcolor}{\textbf{F}}&Na\"ive&$0.99$&4.952\scriptsize{(1.38)}&5.731\scriptsize{(1.42)}&6.835\scriptsize{(1.47)}\\
\textcolor{factoredcolor}{\textbf{F}}&AS&$0.99$&4.395\scriptsize{(0.97)}&4.689\scriptsize{(0.79)}&5.948\scriptsize{(0.93)}\\
\textcolor{factoredcolor}{\textbf{F}}&Na\"ive&$0.9998$&4.467\scriptsize{(0.98)}&\textbf{4.472}\scriptsize{(0.63)}&4.932\scriptsize{(0.52)}\\
% \midrule
\textcolor{factoredcolor}{\textbf{F}}&AS&$0.9998$&\textbf{4.318}\scriptsize{(0.83)}&\underline{4.483}\scriptsize{(0.65)}&\textbf{4.814}\scriptsize{(0.42)}\\
        \bottomrule
    \end{tabular}%
    %\end{adjustbox}%
\end{table*}%

We first evaluate the impact of the three DDQN's design choices described in~\Cref{sec:DQN_with_combinatorial_actions} and~\Cref{sec:DQN_with_factored_actions}, including: (i)  action space representation (combinatorial vs. factored); (ii) reward function (naive vs. adaptive reward shifting, as in~\citet{nguyen2025importance}); and (iii) discount factor ($\gamma \in \{0.99, 0.9998\}$). 

\Cref{tab:compare_rl_reward_gamma} presents the performance of all combinations of those three design choices on three problem sizes $n \in \{100, 200, 500\}$. Interestingly, when the naive reward function and the commonly used discount factor $\gamma=0.99$ are used, there is no clear advantage in using factored action space representation. However, when adaptive reward shifting is adopted, the factored representation starts outperforming its combinatorial counterpart on 2 (/3) problem sizes. When the discount factor is also adjusted to reflect the long episode-horizon characteristic of the benchmark (the last row in \Cref{tab:compare_rl_reward_gamma}), the factored representation shows a clear overall advantage over the combinatorial one: for the problem sizes of $n=100$ and $n=500$, the learned policies are statistically significantly better than all the rest, while for $n=200$, it is not statistically significantly different from the best combination. 

Our results not only confirm the advantage of the factored action space representation, but also highlight the strong interaction between this design choice and the other two components (reward shifting and discount factor adjustment). This observation indicates the intricacy of applying deep-RL in DAC contexts, underscoring the importance of jointly optimizing architectural and algorithmic choices when designing effective deep RL solutions for DAC.

\begin{table*}[ht]
    \centering
    \caption{ERT (and its standard deviation) normalized by the problem size $n$ across 1000 seeds of the best policies found by our DDQN-based multi-parameter control approach and of several baselines from the litearture.}
    \label{tab:compare_all_methods}
    \begin{adjustbox}{max width=0.95\textwidth} % Scale the box to fit the page width
    \begin{tabular}{lccccc}
        \toprule
        % \multirow{2}{*}{\textbf{Problem Size}} & \multicolumn{4}{c}{\textbf{Method}} \\
         % & \multicolumn{6}{c}{\textbf{ERT}($\downarrow$)} \\
        % \cmidrule(lr){2-7}
        & $\pi_{\textsc{theory}}$~\cite{doerr2015black,doerr2018optimal} & \irace~\cite{dang2019hyper} & $1/5$-th~\cite{doerr2015black}&$\pi_{\lambda}$~\cite{nguyen2025importance}&$\pi_{\textsc{mp}}$ \\
        \midrule
        $n=100$ & 5.826\scriptsize{(1.18)}&4.990\scriptsize{(0.86)}&6.197\scriptsize{(1.29)}&5.442\scriptsize{(1.23)}&\textbf{4.318}\scriptsize{(0.83)}\\
        $n=200$ & 6.167\scriptsize{(0.97)}&5.216\scriptsize{(0.69)}&6.541\scriptsize{(1.09)}&5.711\scriptsize{(0.96)}&\textbf{4.483}\scriptsize{(0.65)}\\
        $n=500$ & 6.474\scriptsize{(0.67)}&5.478\scriptsize{(0.46)}&6.744\scriptsize{(0.70)}&6.017\scriptsize{(0.63)}&\textbf{4.814}\scriptsize{(0.42)}\\
        $n=1000$ & 6.587\scriptsize{(0.53)}&5.587\scriptsize{(0.35)}&6.886\scriptsize{(0.54)}&6.338\scriptsize{(0.55)}&\textbf{5.397}\scriptsize{(0.41)}\\
        $n=1500$ & 6.647\scriptsize{(0.44)}&5.621\scriptsize{(0.29)}&6.931\scriptsize{(0.48)}&6.209\scriptsize{(0.41)}&\textbf{4.971}\scriptsize{(0.29)}\\ 
        $n=2000$ & 6.681\scriptsize{(0.39)}&5.666\scriptsize{(0.26)}&7.008\scriptsize{(0.41)}&6.608\scriptsize{(0.41)}&\textbf{5.162}\scriptsize{(0.29)}\\ 

        \bottomrule
    \end{tabular}%
    \end{adjustbox}%
\end{table*} %

Having established that the combination of factored representation, adaptive reward shifting, and discount factor adjustment is the best overall setting for DDQN, we conduct a new set of experiments on a larger set of problem sizes $n \in \{100, 200, 500, 1000, 1500, 2000\}$ using the chosen DDQN setting. In~\Cref{tab:compare_all_methods}, we compare the learned policies obtained from our DDQN-based approach, denoted as $\pi_{\textsc{mp}}$, with several baselines from the literature as described in~\Cref{sec:single-param} and~\Cref{sec:multi-param}, including:
\begin{itemize}
    \item the theory-derived single-parameter control policy proposed in~\cite{doerr2015black,doerr2018optimal}, namely $\pi_{\textsc{theory}}$;
    \item the single-parameter control policy based on the one-fifth success rule~\cite{kern2004learning,auger2009benchmarking};
    \item the multi-parameter control policy obtained from \textsc{irace}~\cite{dang2019hyper};
    \item the DDQN-based single-parameter control policies obtaines in~\cite{nguyen2025importance}, namely $\pi_{\lambda}$.\footnote{The ERT reported in~\cite{nguyen2025importance} was based on average performance across $5$ RL training runs. To ensure a fair comparison, we re-run their experiments and report performance of the best policy across $5$ RL training runs instead of taking the average.}
\end{itemize}

Note that we do not explicitly compare to the results presented in~\cite{AntipovBD22} and~\cite{AntipovBD24} since both of them are much worse than those obtained by $\pi_{\textsc{theory}}$, with ERTs exceeding $12n$ for~\cite{AntipovBD22} and results worse than the (1+1)~EA (and hence much worse than $\pi_{\textsc{theory}}$) in~\cite{AntipovBD24}.

Among the baselines, the \textsc{irace}--based approach~\cite{dang2019hyper} is the strongest one. This is not too surprising, as this is the only baseline that tunes all four parameters of \onell separately. Nevertheless, our multi-parameter control DDQN-based approach statistically significantly outperforms all baselines, including the \textsc{irace}-based one, across all tested problem sizes. 

Our results strongly confirm the advantage of multi-parameter control of \onell on \onemax. In the subsequent sections, we will leverage the obtained results to derive a simple yet powerful control policy for this benchmark.

\begin{figure}[t]
    \centering
    % \begin{subfigure}{0.5\textwidth}
        \includegraphics[width=0.8\linewidth, clip]{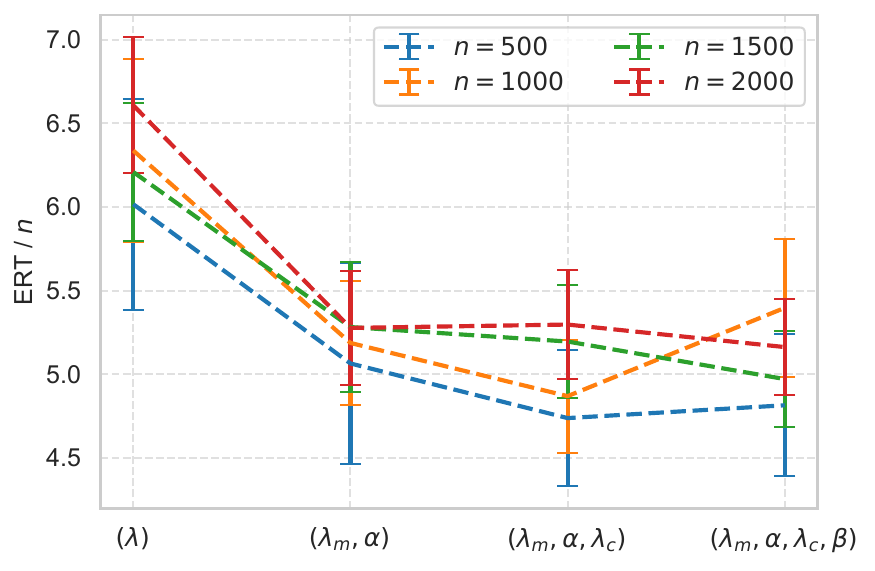}
    \caption{Ablation path from controlling only one parameter of \onell on \onemax to controlling all four parameters with DDQN across different problem sizes. Each point shows the normalized ERTs over 1,000 independent runs and the standard deviation of the corresponding learned policy.}
    \label{fig:ablation_replacement_retraining}
\end{figure}

\section{Ablation Study}
\label{sec:ablation_study}

\begin{table*}[ht]
    \centering
\caption{ERT (and its standard deviation) normalized by the problem size $n$ across 1000 seeds of different versions with varying numbers of controlled parameters. The term \textsc{RL} denotes dynamic control achieved through reinforcement learning, whereas $\alpha=\beta=1$ and $\lambda_c=\lambda_m$ indicate no dynamic control.}
    \label{tab:ablation_important_params}
    %\begin{adjustbox}{max width=0.475\textwidth} % Scale the box to fit the page width
    \begin{tabular}{cccccccc}
        \toprule
         % \multicolumn{3}{c}{\textbf{Setting}} & \multicolumn{3}{c}{\textbf{Problem Size}} \\
        % \midrule 
        $\lambda_m$ & $\alpha$ & $\lambda_c$ & $\beta$ & $\mathbf{n=500}$ & $\mathbf{n=1000}$ & $\mathbf{n=1500}$ & $\mathbf{n=2000}$ \\
        \midrule 
        \multicolumn{4}{c}{$\pi_{\textsc{theory}}$~\cite{doerr2015black,doerr2018optimal}}& 6.474\scriptsize{(0.67)}&6.587\scriptsize{(0.53)}&6.647\scriptsize{(0.44)}&6.681\scriptsize{(0.39)} \\
        \midrule
        \textsc{rl} & 1 & $\lambda_m$ & 1 & 6.017\scriptsize{(0.63)}&6.338\scriptsize{(0.55)}&6.209\scriptsize{(0.41)}&6.608\scriptsize{(0.41)} \\
        % \midrule
        \arrayrulecolor{lightgray}\midrule
        \textsc{rl}& 1 & \textsc{rl} & 1 &5.052\scriptsize{(0.48)}&5.268\scriptsize{(0.38)}&5.504\scriptsize{(0.39)}&6.133\scriptsize{(0.39)}\\
        \textsc{rl}& \textsc{rl} & $\lambda_m$ & 1 &5.064\scriptsize{(0.60)}&5.187\scriptsize{(0.37)}&5.282\scriptsize{(0.39)}&5.277\scriptsize{(0.34)}\\
        \textsc{rl}& 1 & $\lambda_m$ & \textsc{rl} &6.170\scriptsize{(0.67)}&6.220\scriptsize{(0.53)}&6.384\scriptsize{(0.44)}&6.722\scriptsize{(0.44)}\\
        \midrule
        \textsc{rl}& \textsc{rl} & $\lambda_m$ & \textsc{rl}&5.191\scriptsize{(0.63)}&5.303\scriptsize{(0.43)}&5.609\scriptsize{(0.33)}&5.410\scriptsize{(0.32)}\\
        \textsc{rl}& \textsc{rl} & \textsc{rl} & 1 &4.738\scriptsize{(0.40)}&\textbf{4.869}\scriptsize{(0.34)}&5.196\scriptsize{(0.34)}&5.297\scriptsize{(0.32)}\\
        % $(\lambda_m,\lambda_c,\beta)$&2735.568\scriptsize{($\textcolor{blue}{\downarrow16.38\%}$)}&5681.864\scriptsize{($\textcolor{blue}{\downarrow15.21\%}$)}&9056.251\scriptsize{($\textcolor{blue}{\downarrow10.39\%}$)}&12577.681\scriptsize{($\textcolor{blue}{\downarrow7.80\%}$)}\\
        \midrule
         \textsc{rl} &  \textsc{rl} &  \textsc{rl} &  \textsc{rl} &\textbf{4.814}\scriptsize{(0.42)}&\underline{5.397}\scriptsize{(0.41)}&\textbf{4.971}\scriptsize{(0.29)}&\textbf{5.162}\scriptsize{(0.29)}\\

        \arrayrulecolor{black}\bottomrule
    \end{tabular}%
    %\end{adjustbox}%
\end{table*}%

Since our proposed method outperforms others by controlling multiple parameters, we conduct ablation analyses to identify the significance of each parameter. We train the DDQN model with the best setting -- using the factored action space, adaptive reward shifting, and setting the discount factor $\gamma=0.9998$ -- on four problem sizes: $\{500, 1000, 1500, 2000\}$. 

The ablation starts with training DDQN to control only one parameter, which is the mutation population size ($\lambda_m$), while the remaining parameters ($\alpha,\lambda_c,\beta$) are determined by the theory-derived policy $\pi_{\textsc{theory}}$, i.e., $\lambda_c=\lambda_m$ and $\alpha=\beta=1$. Gradually, we integrate more parameters into the set of controllable parameters by RL. For instance, in the second iteration of the ablation, we conduct RL trainings for three combinations: ($\lambda_m$,$\lambda_c$), ($\lambda_m$,$\alpha$), and ($\lambda_m$,$\beta$). We then select the most effective combination as the starting point for the third step, and so on. This is repeated until all four parameters are included. 

Table~\ref{tab:ablation_important_params} shows the performance of all combinations in each ablation iteration across the four studied problem sizes. 

A summary version of Table~\ref{tab:ablation_important_params} where only the best combination at each iteration is shown is presented in~\Cref{fig:ablation_replacement_retraining}.

\begin{figure*}[t]
    \centering
    \includegraphics[width=\linewidth, clip]{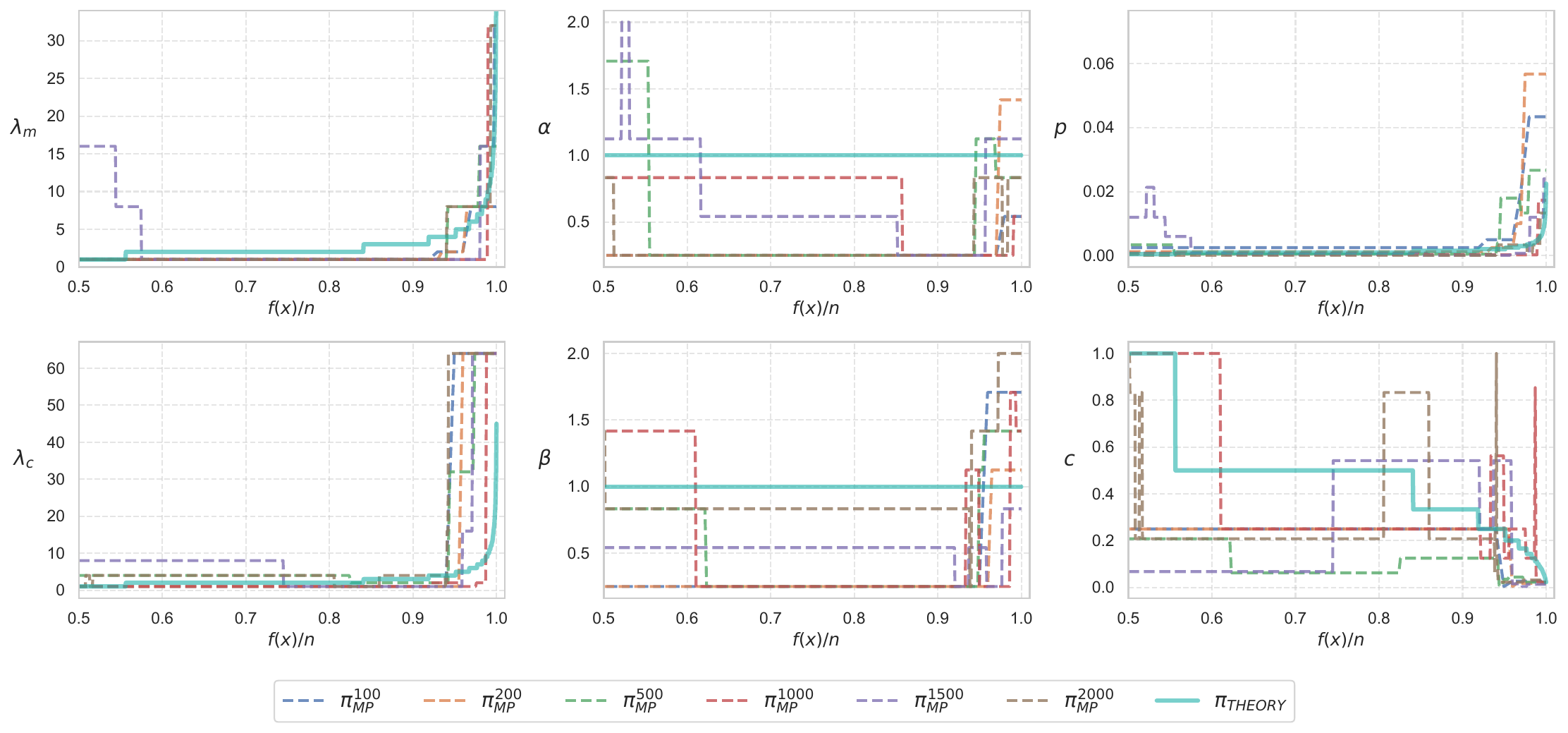}
    %\caption{Merged policies of RL-based MP-DAC across six problem sizes.}
    \caption{DDQN-based policies and the theory-derived policy across six problem sizes of $n \in \{100,200,500,1000,1500,2000\}$. Note that the evolution path of $\lambda_m$ and $\lambda_c$ in the theory-derived policy $\pi_{\textsc{theory}}$ is exactly the same for all problem sizes, however, the upper bound for those parameters increases as problem size increases. Here we plot those values to the maximum upper bound, which corresponds to $n=2000$. In addition to showing the four parameters being controlled by RL, we also show the corresponding mutation rate $p$ and crossover bias $c$ values (third column).}
    \label{fig:merged_policies}
\end{figure*}

When comparing the theory-derived policy and the policies obtained from simultaneously controlling mutation population size ($\lambda_m$) and the mutation rate coefficient ($\alpha$), the average improvements achieved are over $18\%$. Although the average expected runtimes of $(\lambda_m,\alpha)$ do not always outperform $(\lambda_m,\lambda_c)$, our Wilcoxon signed-rank test's results suggest that there is no statistical difference between these two combinations. This observation highlights the crucial role of $\alpha$ in almost four tested problem sizes, and we conclude that $\alpha$ is the next parameter that should be controlled. Therefore, we then verify the impact of integrating one of the two parameters of the crossover phase into the combination. The advancements increase over $22\%$ when we incorporate the offspring population size of crossover $(\lambda_c)$. On the contrary, replacing $\lambda_c$ with $\beta$ adversely affects the overall performance. Controlling $(\lambda_m,\beta)$ also leads to the same observations, suggesting that dynamically adjusting the factor of crossover bias $(\beta)$ is not an effective way to enhance the optimization process. 

The common trend across four problem sizes is shown in~\Cref{fig:ablation_replacement_retraining}, where we see an overall improvement in most cases when a new parameter is added into the set of controllable parameters. There are only a few exceptions, such as the last point in the ablation path of $n=1000$ where the performance got worse. We attribute those cases to DDQN's learning instability and the challenges when the size of the action space is increased. This observation is not too surprising as this kind of learning instability behavior has been observed in previous work on theory-derived dynamic algorithm configuration benchmarks~\cite{biedenkapp2022theory,nguyen2025importance}. We leave for future work a more thorough investigation into improving the learning stability in such cases. In overall, we observe intense plummets during the transition from controlling $\lambda_m$ alone to $(\lambda_m,\alpha)$, followed by $(\lambda_m,\alpha,\lambda_c)$, and finally, we achieve full control over four parameters. This order reflects the important degree of each parameter: $\lambda_m \gg \alpha \gg \lambda_c \gg \beta$. 

\section{Derived Multi-parameter Control Policy}
\label{sec:derived_mp_policy}

Although the policies learned via RL shows strong performance, they are problem-dependent. Moreover, as a neural network-based approach, each learned policy functions as a black box. Those aspects may hinder the usefulness of our findings for theoretical insights. To address this limitation, in this section, based on our ablation analysis and the policies suggested by DDQN, we derive a new symbolic policy, namely $\pi_{\textsc{dmp}}$, that is not only highly interpretable but also exhibits strong performance and generalization. As shown later in this section, our newly derived policy consistently outperforms all existing baselines even on very large problem sizes. We hope this new policy serves as a foundation for future theoretical work on multi-parameter control for \onell on \onemax. 

\begin{figure*}[t]
    \centering
    \includegraphics[width=\linewidth, clip]{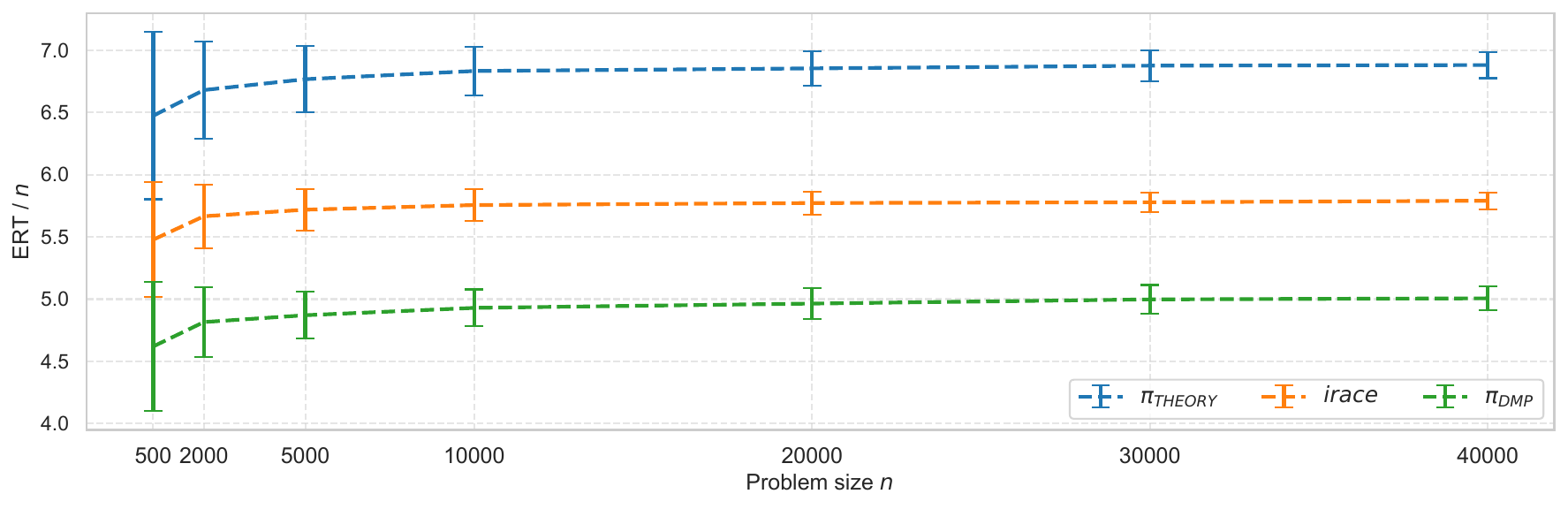}
    %\caption{Merged policies of RL-based MP-DAC across six problem sizes.}
    \caption{ERT (and its standard deviation) normalized by the problem size $n$ across 1000 seeds of our proposed derived policy $\pi_{\textsc{dmp}}$, compared to the theory-derived policy~\cite{doerr2015black,doerr2018optimal} and the \irace-based approach in~\cite{dang2019hyper}.}
    \label{fig:compare_derived_policy_seen_problem_sizes}
\end{figure*}

\begin{table*}[ht]
    \centering
\caption{ERT (and its standard deviation) normalized by the problem size $n$ across 1000 seeds for seven problem sizes of different versions of the derived multi-parameter control policy. 
The final combination of $\pi_{\textsc{dmp}}$ is represented by the last row. Each term in Eq.$(\cdot)$ corresponds to the equation of the new symbolic form for each parameter, while the reference to~\cite{doerr2015black,doerr2018optimal} refers to the theory formula $\lambda_m=\sqrt{n/(n-f(x))}$. The cases where $\alpha=\beta=1$ and $\lambda_c=\lambda_m$ denote no dynamic control.}
    \label{tab:mpdac_derived_policies}
    \begin{adjustbox}{max width=0.99\textwidth} % Scale the box to fit the page width
    \begin{tabular}{ccccccccccc}
        \toprule
        $\lambda_m$ & $\alpha$ &$\lambda_c$ & $\beta$ & ${n=3{,}000}$ & ${n=4{,}000}$ & ${n=5{,}000}$ & ${n=10{,}000}$& ${n=20{,}000}$& ${n=30{,}000}$& ${n=40{,}000}$ \\
        \midrule 
        % $\pi_{\texttt{cont}}$&20167.775\scriptsize{(968.72)}&27050.132\scriptsize{(1111.09)}&33840.312\scriptsize{(1323.31)}&68341.641\scriptsize{(1974.88)} & & & \\
        % \multicolumn{4}{c}{\irace}&17075.429\scriptsize{(641.01)}&22814.871\scriptsize{(774.78)}&28594.172\scriptsize{(837.59)}&57562.200\scriptsize{(1272.34)}&115452.314\scriptsize{(1855.90)}&173359.088\scriptsize{(2327.30)}&231635.664\scriptsize{(2701.22)} \\
        % \midrule
        % $\lambda_m$ & $\alpha$ &$\lambda_c$ & $\beta$ & & & & & & \\
        % \cline{1-4}
        % \cmidrule(lr){1-4}
        \multicolumn{4}{c}{$\pi_{\textsc{theory}}$~\cite{doerr2015black,doerr2018optimal}}&6.723\scriptsize{(0.32)}&6.763\scriptsize{(0.28)}&6.768\scriptsize{(0.26)}&6.834\scriptsize{(0.20)}&6.855\scriptsize{(0.14)}&6.877\scriptsize{(0.13)}&6.882\scriptsize{(0.11)}\\
        \midrule
        Eq.(\ref{eq:new_lbdm})& 1 & $\lambda_m$ & 1 & 6.353\scriptsize{(0.35)}&6.384\scriptsize{(0.30)}&6.404\scriptsize{(0.28)}&6.475\scriptsize{(0.21)}&6.545\scriptsize{(0.16)}&6.581\scriptsize{(0.15)}&6.602\scriptsize{(0.14)}\\
        % \midrule
        % \arrayrulecolor{lightgray}\midrule
        ~\cite{doerr2015black,doerr2018optimal}& Eq.(\ref{eq:new_alpha}) & $\lambda_m$ & 1 &5.905\scriptsize{(0.32)}&5.957\scriptsize{(0.30)}&5.980\scriptsize{(0.26)}&6.044\scriptsize{(0.20)}&6.096\scriptsize{(0.16)}&6.142\scriptsize{(0.15)}&6.157\scriptsize{(0.14)}\\
        % \arrayrulecolor{lightgray}\midrule
        % \textsc{Type I}&17033.113\scriptsize{(699.44)}&22789.210\scriptsize{(819.65)}&28596.153\scriptsize{(985.71)}&57601.892\scriptsize{(1447.51)} \\
        % \textsc{Type II}&16989.753\scriptsize{(730.17)}&22749.724\scriptsize{(877.29)}&28570.843\scriptsize{(1024.20)}&57561.821\scriptsize{(1571.21)} \\
        ~\cite{doerr2015black,doerr2018optimal}& 1 & Eq.(\ref{eq:new_lbdc}) & 1 &5.891\scriptsize{(0.24)}&5.923\scriptsize{(0.22)}&5.930\scriptsize{(0.20)}&5.978\scriptsize{(0.15)}&6.029\scriptsize{(0.12)}&6.051\scriptsize{(0.11)}&6.076\scriptsize{(0.10)}\\

        \arrayrulecolor{lightgray}\midrule
        ~\cite{doerr2015black,doerr2018optimal}& Eq.(\ref{eq:new_alpha}) & Eq.(\ref{eq:new_lbdc}) & 1 &5.167\scriptsize{(0.24)}&5.199\scriptsize{(0.21)}&5.216\scriptsize{(0.19)}&5.269\scriptsize{(0.14)}&5.309\scriptsize{(0.12)}&5.336\scriptsize{(0.11)}&5.362\scriptsize{(0.10)}\\
        % \arrayrulecolor{lightgray}\midrule
        Eq.(\ref{eq:new_lbdm})& 1 & Eq.(\ref{eq:new_lbdc}) & 1 &6.006\scriptsize{(0.26)}&6.022\scriptsize{(0.23)}&6.045\scriptsize{(0.20)}&6.090\scriptsize{(0.16)}&6.132\scriptsize{(0.13)}&6.156\scriptsize{(0.11)}&6.172\scriptsize{(0.10)}\\
        \arrayrulecolor{lightgray}\midrule
        Eq.(\ref{eq:new_lbdm})& Eq.(\ref{eq:new_alpha}) & Eq.(\ref{eq:new_lbdc}) & 1 &\textbf{4.827}\scriptsize{(0.23)}&\textbf{4.859}\scriptsize{(0.21)}&\textbf{4.870}\scriptsize{(0.19)}&\textbf{4.930}\scriptsize{(0.15)}&\textbf{4.964}\scriptsize{(0.12)}&\textbf{4.998}\scriptsize{(0.12)}&\textbf{5.006}\scriptsize{(0.10)}\\
        \arrayrulecolor{black}\bottomrule
    \end{tabular}%
    \end{adjustbox}%
\end{table*}%

\textbf{Definition and Justification for our new policy $\pi_{\textsc{dmp}}$.} 
To derive a new policy, we visualize the DDQN-based policies across all problem sizes based on the normalized fitness values, together with the theory-derived policy, in~\Cref{fig:merged_policies}. 
We first look at $\lambda_m$, an important parameter as indicated in our ablation study in~\Cref{sec:ablation_study}. In general, this parameter is set as one by DDQN for most of the fitness range. Only later in the search, where the fitness value is around $0.95n$ on average, that $\lambda_m$ starts increasing. The magnitudes of $\lambda_m$ in that later part is fairly close to the values suggested by theory (i.e., $\lambda_m=\sqrt{n/(n-f(x))}$).

The next most important parameter according to the ablation is $\alpha$, the mutation rate coefficient. A common trend for this parameter is that it starts jumping to a high value when $f(x) \geq 0.95n$. Therefore, we design a policy where $\alpha$ is set as a small value ($0.001$) during the first part of the search,\footnote{Note that in~\Cref{fig:merged_policies} for a few problem sizes, DDQN appears to suggest a ``U-shape'' policy where $\alpha$ receives a large value before dropping and then increasing. We also experimented with this U-shape policy where we set $\alpha=0.5$ if $f(x) < 0.85n$, $\alpha=0.001$ if $0.85n \leq f(x) \leq 0.95n$, and $\alpha=1$ otherwise. However, this policy performs worse than the one where we simply set $\alpha=0.001$ for all $f(x) \leq 0.95n$. The differences are small, though, this alternative U-shaped policy achieves a normalized ERT of around $6.212$ averaged across $7$ tested problem sizes.}, and is then increased to $1$ when $f(x)$ reaches $0.95n$.

The third parameter we look at is $\lambda_c$, the population size for the crossover phase. Notably, although the theory-derived policy suggests $\lambda_m=\lambda_c$, our DDQN approach imposes a much larger value for $\lambda_c$. The range of this parameter is approximately twice the one suggested by theory. This is similar to the findings in~\cite{dang2019hyper}, where \textsc{irace} suggested that $\lambda_c = 2\lambda_m$. Our DDQN-based approach is able to recognize this correlation without having to explicitly formulate the relationship between $\lambda_c$ and $\lambda_m$. This allows for more flexibility compared to the \textsc{irace}-based approach: since $\lambda_m$ and $\lambda_c$ are completely decoupled from each other, DDQN can choose to ``delay'' increasing $\lambda_m$ during the early part of the search. As we will show in the ablation analysis at the end of this section (\Cref{tab:mpdac_derived_policies}), this design choice has a clear positive contribution to the performance difference between our derived policy and the one found by \textsc{irace}.

Finally, we look at $\beta$, the crossover bias coefficient. The values learned by DDQN for this parameter show a less clear trend, and since this parameter was indicated in our ablation in~\Cref{sec:ablation_study} as the least important one among the four parameters, we decide to keep this parameter as a constant of $1$, i.e., similar to the theory-derived policy.

In summary, we derive a new multi-parameter control policy $\pi_{\textsc{dmp}}$ as follows:

\begin{itemize}
    \item Population size for the mutation phase: 
        \begin{equation}
            \label{eq:new_lbdm}
            \lambda_m =
            \begin{cases}
            1 & \text{if } f(x)/n \le 0.95 \\
            \sqrt{n/(n-f(x))}   & \text{otherwise}
            \end{cases}
        \end{equation}
    \item Population size for the crossover phase: 
        \begin{equation}
        \label{eq:new_lbdc}
        \lambda_c = 2\sqrt{n/(n-f(x))}
        \end{equation}
    \item Mutation rate coefficient:
        \begin{equation}
        \label{eq:new_alpha}
        \alpha =
        \begin{cases}
        0.001 & \text{if } f(x)/n \le 0.95 \\
        1   & \text{otherwise}
        \end{cases}
        \end{equation}
    \item Crossover bias coefficient: $\beta=1$
\end{itemize}

\textbf{Empirical evaluation of $\pi_{\textsc{dmp}}$.} 
We evaluate the newly derived policy across a range of problem sizes from $n=500$ to $n=40{,}000$ and compare against both the theory-derived policy and the multi-parameter control policy found by \textsc{irace}~\cite{dang2019hyper} (the strongest baseline on this benchmark). As shown in~\Cref{fig:compare_derived_policy_seen_problem_sizes}, our policy significantly outperforms the others across all tested problem sizes. Interestingly, the normalized ERT appears to be constant across all problem sizes: it is roughly $6.9$, $5.8$ and $5.0$ for theory-derived policy, the \textsc{irace}-based one, and our DDQN-derived policy, respectively. 

\textbf{Ablation study for $\pi_{\textsc{dmp}}$.} 
To confirm the importance of each component of the newly derived policy, we conduct another ablation where we replace each of the three parameters, $\lambda_m$, $\alpha$, and $\lambda_c$, with the theory-derived policy. The results are shown in~\Cref{tab:mpdac_derived_policies}, where the expected runtime values are again normalized by the problem size. 

We see that $\lambda_c$ plays a crucial role in enhancing ERT, resulting in an average improvement of over $12\%$, followed by the  mutation rate coefficient $\alpha$($\approx 11.5\%$). We next examine the combination of two parameters, where the set of $(\alpha,\lambda_c)$ exhibits an average improvement of over $22.5\%$, significantly outperforming the set of $(\lambda_m,\lambda_c)$ (which only yields a $10\%$ improvement). This observation highlights the negative impact of merging $\lambda_m$ into $\lambda_c$. On the other hand,  $\pi_{\textsc{dmp}}$ that includes a tuple of ($\lambda_m,\alpha,\lambda_c$) precisely captures the contribution of each designed factor. The gap percentage is approximately the sum of each individual gap. Finally, we see that every single parameter in the ablation path has a positive impact on the performance. This observation confirms the importance of all three design choices for $\lambda_m$, $\alpha$, and $\lambda_c$ in our newly derived policy, which outperforms $\pi_{\textsc{theory}}$ by 27.5\%.

\section{Conclusion}

In this work, we have shown that (deep) RL can be a powerful tool to discover high-performing multi-parameter control policies for the \onell on \onemax. Based on results suggested by deep RL, we are able to derive a parameter control policy that outperforms all existing policies on this task. 

As mentioned above, we believe our work to provide new insight for researchers with an interest in reinforcement learning or algorithm configuration and for researchers interested in theoretical aspects of evolutionary computation. For the latter, it is worth noting that our derived policy $\pi_{\textsc{dmp}}$ is fairly simple and may be tractable for a fine-grained running time analysis. However, we note that, to date, no reasonably good bounds for the constant factors of the linear expected running time of the \onell with suitable dynamic parameter choices are known, even though experimental results for up to $n=2^{22}$ in~\citet{AntipovBD22} suggest that the constant factor of the default 1/5-th success rule proposed in~\cite{doerr2015black,doerr2018optimal} is somewhere around $6.9$, perfectly in line with our empirical results. The interested may have noticed that the constant factors for all algorithms and parameter control policies increase with increasing problem dimension $n$, and this includes the settings for which we have proven guarantees of linear running times. However, as~\citet[Section~4.3]{AntipovBD22} acknowledges, the few formally proven upper bounds for the constant factors (for their algorithm choosing $\lambda$ from a heavy-tailed power-law distribution) seem to be far too pessimistic. Since the precision of running time results has drastically increased over the last years, fine-grained performance analyses of the original self-adjusting \onell and the \onell equipped with the $\pi_{\textsc{dmp}}$ policy may already be feasible or become feasible in the near future. 

From a more general perspective, we hope that this work, in line with other recent works such as~\cite{VermettenLRBD24} (expanding upon~\cite{LenglerR22}) and~\cite{DoerrDL21} (deriving running time guarantees for observations made in~\cite{DoerrW18}) underlines the benefits of tying sound empirical investigations with rigorous theoretical analyses, and vice-versa.  

On the empirical side, in addition to further exploring the DDQN approach used in this work in the context of dynamic algorithm configuration, we plan to conduct an extensive and systematic empirical study into other (deep) RL approaches for multi-parameter control, such as multi-agent deep RL~\cite{xue2022multi}. Additionally, we aim to explore the potential performance improvements that could be achieved by learning in continuous action spaces. This would enable more fine-grained control over the four parameters but would necessitate adopting a different class of deep RL algorithms, such as Proximal Policy Optimization (PPO)~\cite{schulman2017proximal} or Soft Actor-Critic (SAC)~\cite{HaarnojaZAL18}. A further promising direction for future research involves leveraging symbolic regression techniques to automatically derive interpretable parameter control policies, guided by the insights obtained from deep RL models.

\begin{acks}
The project is financially supported by the European Union (ERC, ``dynaBBO'', grant no.~101125586), by ANR project ANR-23-CE23-0035 Opt4DAC, and by an International Emerging Action funded by CNRS Sciences informatiques. This work used the supercomputer at MeSU Platform (\href{https://sacado.sorbonne-universite.fr/plateforme-mesu}{https://sacado.sorbonne-universite.fr/plateforme-mesu}). Tai Nguyen acknowledges funding from the St Andrews Global Doctoral Scholarship programme. This publication is based upon work from COST Action CA22137 ``Randomized Optimization Algorithms Research Network'' (ROAR-NET), supported by COST (European Cooperation in Science and Technology).
\end{acks}

% \newpage
\bibliographystyle{ACM-Reference-Format}
\bibliography{refs}


\begin{thebibliography}{59}

%%% the Web address.

\ifx \showCODEN    \undefined \def \showCODEN     #1{\unskip}     \fi
\ifx \showISBNx    \undefined \def \showISBNx     #1{\unskip}     \fi
\ifx \showISBNxiii \undefined \def \showISBNxiii  #1{\unskip}     \fi
\ifx \showISSN     \undefined \def \showISSN      #1{\unskip}     \fi
\ifx \showLCCN     \undefined \def \showLCCN      #1{\unskip}     \fi
\ifx \shownote     \undefined \def \shownote      #1{#1}          \fi
\ifx \showarticletitle \undefined \def \showarticletitle #1{#1}   \fi
\ifx \showURL      \undefined \def \showURL       {\relax}        \fi
% invisible to TeX
\providecommand\bibfield[2]{#2}
\providecommand\bibinfo[2]{#2}
\providecommand\natexlab[1]{#1}
\providecommand\showeprint[2][]{arXiv:#2}

\bibitem[Adriaensen et~al\mbox{.}(2022)]%
        {adriaensen2022automated}
\bibfield{author}{\bibinfo{person}{Steven Adriaensen}, \bibinfo{person}{Andr{\'e} Biedenkapp}, \bibinfo{person}{Gresa Shala}, \bibinfo{person}{Noor Awad}, \bibinfo{person}{Theresa Eimer}, \bibinfo{person}{Marius Lindauer}, {and} \bibinfo{person}{Frank Hutter}.} \bibinfo{year}{2022}\natexlab{}.
\newblock \showarticletitle{Automated dynamic algorithm configuration}.
\newblock \bibinfo{journal}{\emph{Journal of Artificial Intelligence Research}}  \bibinfo{volume}{75} (\bibinfo{year}{2022}), \bibinfo{pages}{1633--1699}.
\newblock


\bibitem[Aleti and Moser(2016)]%
        {AletiM16}
\bibfield{author}{\bibinfo{person}{Aldeida Aleti} {and} \bibinfo{person}{Irene Moser}.} \bibinfo{year}{2016}\natexlab{}.
\newblock \showarticletitle{A Systematic Literature Review of Adaptive Parameter Control Methods for Evolutionary Algorithms}.
\newblock \bibinfo{journal}{\emph{Comput. Surveys}}  \bibinfo{volume}{49} (\bibinfo{year}{2016}), \bibinfo{pages}{56:1--56:35}.
\newblock


\bibitem[Antipov et~al\mbox{.}(2022a)]%
        {AntipovBD22}
\bibfield{author}{\bibinfo{person}{Denis Antipov}, \bibinfo{person}{Maxim Buzdalov}, {and} \bibinfo{person}{Benjamin Doerr}.} \bibinfo{year}{2022}\natexlab{a}.
\newblock \showarticletitle{Fast Mutation in Crossover-Based Algorithms}.
\newblock \bibinfo{journal}{\emph{Algorithmica}} \bibinfo{volume}{84}, \bibinfo{number}{6} (\bibinfo{year}{2022}), \bibinfo{pages}{1724--1761}.
\newblock
\href{https://doi.org/10.1007/S00453-022-00957-5}{doi:\nolinkurl{10.1007/S00453-022-00957-5}}


\bibitem[Antipov et~al\mbox{.}(2024)]%
        {AntipovBD24}
\bibfield{author}{\bibinfo{person}{Denis Antipov}, \bibinfo{person}{Maxim Buzdalov}, {and} \bibinfo{person}{Benjamin Doerr}.} \bibinfo{year}{2024}\natexlab{}.
\newblock \showarticletitle{Lazy Parameter Tuning and Control: Choosing All Parameters Randomly from a Power-Law Distribution}.
\newblock \bibinfo{journal}{\emph{Algorithmica}} \bibinfo{volume}{86}, \bibinfo{number}{2} (\bibinfo{year}{2024}), \bibinfo{pages}{442--484}.
\newblock
\href{https://doi.org/10.1007/S00453-023-01098-Z}{doi:\nolinkurl{10.1007/S00453-023-01098-Z}}


\bibitem[Antipov et~al\mbox{.}(2019)]%
        {AntipovDK19}
\bibfield{author}{\bibinfo{person}{Denis Antipov}, \bibinfo{person}{Benjamin Doerr}, {and} \bibinfo{person}{Vitalii Karavaev}.} \bibinfo{year}{2019}\natexlab{}.
\newblock \showarticletitle{A tight runtime analysis for the {(1} + ({\(\lambda\)}, {\(\lambda\)})) {GA} on {L}eading{O}nes}. In \bibinfo{booktitle}{\emph{Proc. of {ACM/SIGEVO} Conference on Foundations of Genetic Algorithms (FOGA)}}, \bibfield{editor}{\bibinfo{person}{Tobias Friedrich}, \bibinfo{person}{Carola Doerr}, {and} \bibinfo{person}{Dirk~V. Arnold}} (Eds.). \bibinfo{publisher}{{ACM}}, \bibinfo{pages}{169--182}.
\newblock
\href{https://doi.org/10.1145/3299904.3340317}{doi:\nolinkurl{10.1145/3299904.3340317}}


\bibitem[Antipov et~al\mbox{.}(2022b)]%
        {AntipovDK22}
\bibfield{author}{\bibinfo{person}{Denis Antipov}, \bibinfo{person}{Benjamin Doerr}, {and} \bibinfo{person}{Vitalii Karavaev}.} \bibinfo{year}{2022}\natexlab{b}.
\newblock \showarticletitle{A Rigorous Runtime Analysis of the {(1} + ({\(\lambda\)} , {\(\lambda\)} {))} {GA} on Jump Functions}.
\newblock \bibinfo{journal}{\emph{Algorithmica}} \bibinfo{volume}{84}, \bibinfo{number}{6} (\bibinfo{year}{2022}), \bibinfo{pages}{1573--1602}.
\newblock
\href{https://doi.org/10.1007/S00453-021-00907-7}{doi:\nolinkurl{10.1007/S00453-021-00907-7}}


\bibitem[Auger(2009)]%
        {auger2009benchmarking}
\bibfield{author}{\bibinfo{person}{Anne Auger}.} \bibinfo{year}{2009}\natexlab{}.
\newblock \showarticletitle{Benchmarking the (1+ 1) evolution strategy with one-fifth success rule on the BBOB-2009 function testbed}. In \bibinfo{booktitle}{\emph{Proceedings of the 11th Annual Conference Companion on Genetic and Evolutionary Computation Conference: Late Breaking Papers}}. \bibinfo{pages}{2447--2452}.
\newblock


\bibitem[Bassin et~al\mbox{.}(2021)]%
        {BassinBS21}
\bibfield{author}{\bibinfo{person}{Anton~O. Bassin}, \bibinfo{person}{Maxim~V. Buzdalov}, {and} \bibinfo{person}{Anatoly~A. Shalyto}.} \bibinfo{year}{2021}\natexlab{}.
\newblock \showarticletitle{The "One-Fifth Rule" with Rollbacks for Self-Adjustment of the Population Size in the {(1} + ({\(\lambda\)}, {\(\lambda\)})) Genetic Algorithm}.
\newblock \bibinfo{journal}{\emph{Autom. Control. Comput. Sci.}} \bibinfo{volume}{55}, \bibinfo{number}{7} (\bibinfo{year}{2021}), \bibinfo{pages}{885--902}.
\newblock
\href{https://doi.org/10.3103/S0146411621070208}{doi:\nolinkurl{10.3103/S0146411621070208}}


\bibitem[Biedenkapp et~al\mbox{.}(2020)]%
        {biedenkapp2020dynamic}
\bibfield{author}{\bibinfo{person}{Andr{\'e} Biedenkapp}, \bibinfo{person}{H~Furkan Bozkurt}, \bibinfo{person}{Theresa Eimer}, \bibinfo{person}{Frank Hutter}, {and} \bibinfo{person}{Marius Lindauer}.} \bibinfo{year}{2020}\natexlab{}.
\newblock \showarticletitle{Dynamic algorithm configuration: Foundation of a new meta-algorithmic framework}.
\newblock In \bibinfo{booktitle}{\emph{ECAI 2020}}. \bibinfo{publisher}{IOS Press}, \bibinfo{pages}{427--434}.
\newblock


\bibitem[Biedenkapp et~al\mbox{.}(2022)]%
        {biedenkapp2022theory}
\bibfield{author}{\bibinfo{person}{Andr{\'e} Biedenkapp}, \bibinfo{person}{Nguyen Dang}, \bibinfo{person}{Martin~S Krejca}, \bibinfo{person}{Frank Hutter}, {and} \bibinfo{person}{Carola Doerr}.} \bibinfo{year}{2022}\natexlab{}.
\newblock \showarticletitle{Theory-inspired parameter control benchmarks for dynamic algorithm configuration}. In \bibinfo{booktitle}{\emph{Proceedings of the Genetic and Evolutionary Computation Conference}}. \bibinfo{pages}{766--775}.
\newblock


\bibitem[Buzdalov and Doerr(2017)]%
        {BuzdalovD17}
\bibfield{author}{\bibinfo{person}{Maxim Buzdalov} {and} \bibinfo{person}{Benjamin Doerr}.} \bibinfo{year}{2017}\natexlab{}.
\newblock \showarticletitle{Runtime analysis of the {(1} + (\emph{{\(\lambda\)}, {\(\lambda\)}})) genetic algorithm on random satisfiable 3-CNF formulas}. In \bibinfo{booktitle}{\emph{Proce. of Genetic and Evolutionary Computation Conference (GECCO)}}, \bibfield{editor}{\bibinfo{person}{Peter A.~N. Bosman}} (Ed.). \bibinfo{publisher}{{ACM}}, \bibinfo{pages}{1343--1350}.
\newblock
\href{https://doi.org/10.1145/3071178.3071297}{doi:\nolinkurl{10.1145/3071178.3071297}}


\bibitem[Chen et~al\mbox{.}(2023)]%
        {chen2023using}
\bibfield{author}{\bibinfo{person}{Deyao Chen}, \bibinfo{person}{Maxim Buzdalov}, \bibinfo{person}{Carola Doerr}, {and} \bibinfo{person}{Nguyen Dang}.} \bibinfo{year}{2023}\natexlab{}.
\newblock \showarticletitle{Using automated algorithm configuration for parameter control}. In \bibinfo{booktitle}{\emph{Proceedings of the 17th ACM/SIGEVO Conference on Foundations of Genetic Algorithms}}. \bibinfo{pages}{38--49}.
\newblock


\bibitem[Covington et~al\mbox{.}(2016)]%
        {covington2016deep}
\bibfield{author}{\bibinfo{person}{Paul Covington}, \bibinfo{person}{Jay Adams}, {and} \bibinfo{person}{Emre Sargin}.} \bibinfo{year}{2016}\natexlab{}.
\newblock \showarticletitle{Deep neural networks for youtube recommendations}. In \bibinfo{booktitle}{\emph{Proceedings of the 10th ACM conference on recommender systems}}. \bibinfo{pages}{191--198}.
\newblock


\bibitem[Dang and Doerr(2019)]%
        {dang2019hyper}
\bibfield{author}{\bibinfo{person}{Nguyen Dang} {and} \bibinfo{person}{Carola Doerr}.} \bibinfo{year}{2019}\natexlab{}.
\newblock \showarticletitle{Hyper-parameter tuning for the (1+($\lambda$, $\lambda$)) GA}. In \bibinfo{booktitle}{\emph{Proceedings of the Genetic and Evolutionary Computation Conference}}. \bibinfo{pages}{889--897}.
\newblock


\bibitem[Devroye(1972)]%
        {Devroye72}
\bibfield{author}{\bibinfo{person}{Luc Devroye}.} \bibinfo{year}{1972}\natexlab{}.
\newblock \bibinfo{booktitle}{\emph{The compound random search}}.
\newblock \bibinfo{publisher}{Ph.D. dissertation, Purdue Univ., West Lafayette, {IN}}.
\newblock


\bibitem[Doerr and Doerr(2015)]%
        {doerr2015optimal}
\bibfield{author}{\bibinfo{person}{Benjamin Doerr} {and} \bibinfo{person}{Carola Doerr}.} \bibinfo{year}{2015}\natexlab{}.
\newblock \showarticletitle{Optimal parameter choices through self-adjustment: Applying the 1/5-th rule in discrete settings}. In \bibinfo{booktitle}{\emph{Proceedings of the 2015 Annual Conference on Genetic and Evolutionary Computation}}. \bibinfo{pages}{1335--1342}.
\newblock


\bibitem[Doerr and Doerr(2018)]%
        {doerr2018optimal}
\bibfield{author}{\bibinfo{person}{Benjamin Doerr} {and} \bibinfo{person}{Carola Doerr}.} \bibinfo{year}{2018}\natexlab{}.
\newblock \showarticletitle{Optimal static and self-adjusting parameter choices for the (1+($\lambda$, $\lambda$)) genetic algorithm}.
\newblock \bibinfo{journal}{\emph{Algorithmica}}  \bibinfo{volume}{80} (\bibinfo{year}{2018}), \bibinfo{pages}{1658--1709}.
\newblock


\bibitem[Doerr and Doerr(2020)]%
        {doerr2020theory}
\bibfield{author}{\bibinfo{person}{Benjamin Doerr} {and} \bibinfo{person}{Carola Doerr}.} \bibinfo{year}{2020}\natexlab{}.
\newblock \showarticletitle{Theory of parameter control for discrete black-box optimization: Provable performance gains through dynamic parameter choices}.
\newblock \bibinfo{journal}{\emph{Theory of Evolutionary Computation: Recent Developments in Discrete Optimization}} (\bibinfo{year}{2020}), \bibinfo{pages}{271--321}.
\newblock


\bibitem[Doerr et~al\mbox{.}(2015)]%
        {doerr2015black}
\bibfield{author}{\bibinfo{person}{Benjamin Doerr}, \bibinfo{person}{Carola Doerr}, {and} \bibinfo{person}{Franziska Ebel}.} \bibinfo{year}{2015}\natexlab{}.
\newblock \showarticletitle{From black-box complexity to designing new genetic algorithms}.
\newblock \bibinfo{journal}{\emph{Theoretical Computer Science}}  \bibinfo{volume}{567} (\bibinfo{year}{2015}), \bibinfo{pages}{87--104}.
\newblock


\bibitem[Doerr et~al\mbox{.}(2021)]%
        {DoerrDL21}
\bibfield{author}{\bibinfo{person}{Benjamin Doerr}, \bibinfo{person}{Carola Doerr}, {and} \bibinfo{person}{Johannes Lengler}.} \bibinfo{year}{2021}\natexlab{}.
\newblock \showarticletitle{Self-Adjusting Mutation Rates with Provably Optimal Success Rules}.
\newblock \bibinfo{journal}{\emph{Algorithmica}} \bibinfo{volume}{83}, \bibinfo{number}{10} (\bibinfo{year}{2021}), \bibinfo{pages}{3108--3147}.
\newblock
\href{https://doi.org/10.1007/S00453-021-00854-3}{doi:\nolinkurl{10.1007/S00453-021-00854-3}}


\bibitem[Doerr and Winzen(2014)]%
        {doerr2014playing}
\bibfield{author}{\bibinfo{person}{Benjamin Doerr} {and} \bibinfo{person}{Carola Winzen}.} \bibinfo{year}{2014}\natexlab{}.
\newblock \showarticletitle{Playing Mastermind with constant-size memory}.
\newblock \bibinfo{journal}{\emph{Theory of Computing Systems}}  \bibinfo{volume}{55} (\bibinfo{year}{2014}), \bibinfo{pages}{658--684}.
\newblock


\bibitem[Doerr and Wagner(2018)]%
        {DoerrW18}
\bibfield{author}{\bibinfo{person}{Carola Doerr} {and} \bibinfo{person}{Markus Wagner}.} \bibinfo{year}{2018}\natexlab{}.
\newblock \showarticletitle{Simple on-the-fly parameter selection mechanisms for two classical discrete black-box optimization benchmark problems}. In \bibinfo{booktitle}{\emph{Proc. of Genetic and Evolutionary Computation Conference (GECCO)}}, \bibfield{editor}{\bibinfo{person}{Hern{\'{a}}n~E. Aguirre} {and} \bibinfo{person}{Keiki Takadama}} (Eds.). \bibinfo{publisher}{{ACM}}, \bibinfo{pages}{943--950}.
\newblock
\href{https://doi.org/10.1145/3205455.3205560}{doi:\nolinkurl{10.1145/3205455.3205560}}


\bibitem[Dulac-Arnold et~al\mbox{.}({[n.\,d.]})]%
        {dulac1512deep}
\bibfield{author}{\bibinfo{person}{G Dulac-Arnold}, \bibinfo{person}{R Evans}, \bibinfo{person}{H van Hasselt}, \bibinfo{person}{P Sunehag}, \bibinfo{person}{T Lillicrap}, \bibinfo{person}{J Hunt}, \bibinfo{person}{T Mann}, \bibinfo{person}{T Weber}, \bibinfo{person}{T Degris}, {and} \bibinfo{person}{B Coppin}.} \bibinfo{year}{[n.\,d.]}\natexlab{}.
\newblock \showarticletitle{Deep reinforcement learning in large discrete action spaces. arXiv 2015}.
\newblock \bibinfo{journal}{\emph{arXiv preprint arXiv:1512.07679}} (\bibinfo{year}{[n.\,d.]}).
\newblock


\bibitem[Dulac-Arnold et~al\mbox{.}(2019)]%
        {dulac2019challenges}
\bibfield{author}{\bibinfo{person}{Gabriel Dulac-Arnold}, \bibinfo{person}{Daniel Mankowitz}, {and} \bibinfo{person}{Todd Hester}.} \bibinfo{year}{2019}\natexlab{}.
\newblock \showarticletitle{Challenges of real-world reinforcement learning}.
\newblock \bibinfo{journal}{\emph{arXiv preprint arXiv:1904.12901}} (\bibinfo{year}{2019}).
\newblock


\bibitem[Eiben et~al\mbox{.}(1999)]%
        {EibenHM99}
\bibfield{author}{\bibinfo{person}{Agoston~Endre Eiben}, \bibinfo{person}{Robert Hinterding}, {and} \bibinfo{person}{Zbigniew Michalewicz}.} \bibinfo{year}{1999}\natexlab{}.
\newblock \showarticletitle{Parameter control in evolutionary algorithms}.
\newblock \bibinfo{journal}{\emph{IEEE Transactions on Evolutionary Computation}}  \bibinfo{volume}{3} (\bibinfo{year}{1999}), \bibinfo{pages}{124--141}.
\newblock


\bibitem[Goldman and Punch(2015)]%
        {GoldmanP15}
\bibfield{author}{\bibinfo{person}{Brian~W. Goldman} {and} \bibinfo{person}{William~F. Punch}.} \bibinfo{year}{2015}\natexlab{}.
\newblock \showarticletitle{Fast and Efficient Black Box Optimization Using the Parameter-less Population Pyramid}.
\newblock \bibinfo{journal}{\emph{Evolutionary Computation}}  \bibinfo{volume}{23} (\bibinfo{year}{2015}), \bibinfo{pages}{451--479}.
\newblock


\bibitem[Haarnoja et~al\mbox{.}(2018)]%
        {HaarnojaZAL18}
\bibfield{author}{\bibinfo{person}{Tuomas Haarnoja}, \bibinfo{person}{Aurick Zhou}, \bibinfo{person}{Pieter Abbeel}, {and} \bibinfo{person}{Sergey Levine}.} \bibinfo{year}{2018}\natexlab{}.
\newblock \showarticletitle{Soft Actor-Critic: Off-Policy Maximum Entropy Deep Reinforcement Learning with a Stochastic Actor.}. In \bibinfo{booktitle}{\emph{ICML}} \emph{(\bibinfo{series}{Proceedings of Machine Learning Research}, Vol.~\bibinfo{volume}{80})}, \bibfield{editor}{\bibinfo{person}{Jennifer~G. Dy} {and} \bibinfo{person}{Andreas Krause}} (Eds.). \bibinfo{publisher}{PMLR}, \bibinfo{pages}{1856--1865}.
\newblock
\urldef\tempurl%
\url{http://dblp.uni-trier.de/db/conf/icml/icml2018.html#HaarnojaZAL18}
\showURL{%
\tempurl}


\bibitem[He et~al\mbox{.}(2016)]%
        {he-etal-2016-deep-reinforcement}
\bibfield{author}{\bibinfo{person}{Ji He}, \bibinfo{person}{Mari Ostendorf}, \bibinfo{person}{Xiaodong He}, \bibinfo{person}{Jianshu Chen}, \bibinfo{person}{Jianfeng Gao}, \bibinfo{person}{Lihong Li}, {and} \bibinfo{person}{Li Deng}.} \bibinfo{year}{2016}\natexlab{}.
\newblock \showarticletitle{Deep Reinforcement Learning with a Combinatorial Action Space for Predicting Popular {R}eddit Threads}. In \bibinfo{booktitle}{\emph{Proceedings of the 2016 Conference on Empirical Methods in Natural Language Processing}}, \bibfield{editor}{\bibinfo{person}{Jian Su}, \bibinfo{person}{Kevin Duh}, {and} \bibinfo{person}{Xavier Carreras}} (Eds.). \bibinfo{publisher}{Association for Computational Linguistics}, \bibinfo{address}{Austin, Texas}, \bibinfo{pages}{1838--1848}.
\newblock
\href{https://doi.org/10.18653/v1/D16-1189}{doi:\nolinkurl{10.18653/v1/D16-1189}}


\bibitem[Henderson et~al\mbox{.}(2018)]%
        {henderson2018deep}
\bibfield{author}{\bibinfo{person}{Peter Henderson}, \bibinfo{person}{Riashat Islam}, \bibinfo{person}{Philip Bachman}, \bibinfo{person}{Joelle Pineau}, \bibinfo{person}{Doina Precup}, {and} \bibinfo{person}{David Meger}.} \bibinfo{year}{2018}\natexlab{}.
\newblock \showarticletitle{Deep reinforcement learning that matters}. In \bibinfo{booktitle}{\emph{Proceedings of the AAAI conference on artificial intelligence}}, Vol.~\bibinfo{volume}{32}.
\newblock


\bibitem[Hu et~al\mbox{.}(2022)]%
        {hu2022role}
\bibfield{author}{\bibinfo{person}{Hao Hu}, \bibinfo{person}{Yiqin Yang}, \bibinfo{person}{Qianchuan Zhao}, {and} \bibinfo{person}{Chongjie Zhang}.} \bibinfo{year}{2022}\natexlab{}.
\newblock \showarticletitle{On the role of discount factor in offline reinforcement learning}. In \bibinfo{booktitle}{\emph{International conference on machine learning}}. PMLR, \bibinfo{pages}{9072--9098}.
\newblock


\bibitem[Islam et~al\mbox{.}(2017)]%
        {islam2017reproducibility}
\bibfield{author}{\bibinfo{person}{Riashat Islam}, \bibinfo{person}{Peter Henderson}, \bibinfo{person}{Maziar Gomrokchi}, {and} \bibinfo{person}{Doina Precup}.} \bibinfo{year}{2017}\natexlab{}.
\newblock \showarticletitle{Reproducibility of benchmarked deep reinforcement learning tasks for continuous control}.
\newblock \bibinfo{journal}{\emph{arXiv preprint arXiv:1708.04133}} (\bibinfo{year}{2017}).
\newblock


\bibitem[Karafotias et~al\mbox{.}(2012)]%
        {KarafotiasSE12}
\bibfield{author}{\bibinfo{person}{Giorgos Karafotias}, \bibinfo{person}{Selmar~K. Smit}, {and} \bibinfo{person}{A.~E. Eiben}.} \bibinfo{year}{2012}\natexlab{}.
\newblock \showarticletitle{A Generic Approach to Parameter Control}. In \bibinfo{booktitle}{\emph{Proc. of Applications of Evolutionary Computation (EvoApplications'12)}} \emph{(\bibinfo{series}{LNCS}, Vol.~\bibinfo{volume}{7248})}. \bibinfo{publisher}{Springer}, \bibinfo{pages}{366--375}.
\newblock
\href{https://doi.org/10.1007/978-3-642-29178-4\_37}{doi:\nolinkurl{10.1007/978-3-642-29178-4\_37}}


\bibitem[Kern et~al\mbox{.}(2004a)]%
        {KernMHBOK04}
\bibfield{author}{\bibinfo{person}{Stefan Kern}, \bibinfo{person}{Sibylle~D. M{\"{u}}ller}, \bibinfo{person}{Nikolaus Hansen}, \bibinfo{person}{Dirk B{\"{u}}che}, \bibinfo{person}{Jiri Ocenasek}, {and} \bibinfo{person}{Petros Koumoutsakos}.} \bibinfo{year}{2004}\natexlab{a}.
\newblock \showarticletitle{Learning probability distributions in continuous evolutionary algorithms - a comparative review}.
\newblock \bibinfo{journal}{\emph{Natural Computing}}  \bibinfo{volume}{3} (\bibinfo{year}{2004}), \bibinfo{pages}{77--112}.
\newblock


\bibitem[Kern et~al\mbox{.}(2004b)]%
        {kern2004learning}
\bibfield{author}{\bibinfo{person}{Stefan Kern}, \bibinfo{person}{Sibylle~D M{\"u}ller}, \bibinfo{person}{Nikolaus Hansen}, \bibinfo{person}{Dirk B{\"u}che}, \bibinfo{person}{Jiri Ocenasek}, {and} \bibinfo{person}{Petros Koumoutsakos}.} \bibinfo{year}{2004}\natexlab{b}.
\newblock \showarticletitle{Learning probability distributions in continuous evolutionary algorithms--a comparative review}.
\newblock \bibinfo{journal}{\emph{Natural Computing}}  \bibinfo{volume}{3} (\bibinfo{year}{2004}), \bibinfo{pages}{77--112}.
\newblock


\bibitem[Kingma and Ba(2015)]%
        {adam}
\bibfield{author}{\bibinfo{person}{Diederik~P. Kingma} {and} \bibinfo{person}{Jimmy Ba}.} \bibinfo{year}{2015}\natexlab{}.
\newblock \showarticletitle{Adam: {A} Method for Stochastic Optimization}. In \bibinfo{booktitle}{\emph{Proceedings of the 3rd International Conference on Learning Representations, ({ICLR}'15)}}, \bibfield{editor}{\bibinfo{person}{Yoshua Bengio} {and} \bibinfo{person}{Yann LeCun}} (Eds.).
\newblock


\bibitem[Lehre and Witt(2012)]%
        {LehreW12}
\bibfield{author}{\bibinfo{person}{Per~Kristian Lehre} {and} \bibinfo{person}{Carsten Witt}.} \bibinfo{year}{2012}\natexlab{}.
\newblock \showarticletitle{Black-Box Search by Unbiased Variation}.
\newblock \bibinfo{journal}{\emph{Algorithmica}} \bibinfo{volume}{64}, \bibinfo{number}{4} (\bibinfo{year}{2012}), \bibinfo{pages}{623--642}.
\newblock
\href{https://doi.org/10.1007/S00453-012-9616-8}{doi:\nolinkurl{10.1007/S00453-012-9616-8}}


\bibitem[Lengler and Riedi(2022)]%
        {LenglerR22}
\bibfield{author}{\bibinfo{person}{Johannes Lengler} {and} \bibinfo{person}{Simone Riedi}.} \bibinfo{year}{2022}\natexlab{}.
\newblock \showarticletitle{Runtime Analysis of the ({\(\mu\)} + 1)-EA on the Dynamic BinVal Function}.
\newblock \bibinfo{journal}{\emph{{SN} Comput. Sci.}} \bibinfo{volume}{3}, \bibinfo{number}{4} (\bibinfo{year}{2022}), \bibinfo{pages}{324}.
\newblock
\href{https://doi.org/10.1007/S42979-022-01203-Z}{doi:\nolinkurl{10.1007/S42979-022-01203-Z}}


\bibitem[Lillicrap et~al\mbox{.}(2016)]%
        {ddpg-soft-update}
\bibfield{author}{\bibinfo{person}{Timothy~P. Lillicrap}, \bibinfo{person}{Jonathan~J. Hunt}, \bibinfo{person}{Alexander Pritzel}, \bibinfo{person}{Nicolas Heess}, \bibinfo{person}{Tom Erez}, \bibinfo{person}{Yuval Tassa}, \bibinfo{person}{David Silver}, {and} \bibinfo{person}{Daan Wierstra}.} \bibinfo{year}{2016}\natexlab{}.
\newblock \showarticletitle{Continuous control with deep reinforcement learning}. In \bibinfo{booktitle}{\emph{4th International Conference on Learning Representations, {ICLR} 2016, San Juan, Puerto Rico, May 2-4, 2016, Conference Track Proceedings}}, \bibfield{editor}{\bibinfo{person}{Yoshua Bengio} {and} \bibinfo{person}{Yann LeCun}} (Eds.).
\newblock
\urldef\tempurl%
\url{http://arxiv.org/abs/1509.02971}
\showURL{%
\tempurl}


\bibitem[L{\'o}pez-Ib{\'a}{\~n}ez et~al\mbox{.}(2016)]%
        {lopez2016irace}
\bibfield{author}{\bibinfo{person}{Manuel L{\'o}pez-Ib{\'a}{\~n}ez}, \bibinfo{person}{J{\'e}r{\'e}mie Dubois-Lacoste}, \bibinfo{person}{Leslie~P{\'e}rez C{\'a}ceres}, \bibinfo{person}{Mauro Birattari}, {and} \bibinfo{person}{Thomas St{\"u}tzle}.} \bibinfo{year}{2016}\natexlab{}.
\newblock \showarticletitle{The irace package: Iterated racing for automatic algorithm configuration}.
\newblock \bibinfo{journal}{\emph{Operations Research Perspectives}}  \bibinfo{volume}{3} (\bibinfo{year}{2016}), \bibinfo{pages}{43--58}.
\newblock


\bibitem[Ma et~al\mbox{.}(2024)]%
        {ma2024auto}
\bibfield{author}{\bibinfo{person}{Zeyuan Ma}, \bibinfo{person}{Jiacheng Chen}, \bibinfo{person}{Hongshu Guo}, \bibinfo{person}{Yining Ma}, {and} \bibinfo{person}{Yue-Jiao Gong}.} \bibinfo{year}{2024}\natexlab{}.
\newblock \showarticletitle{Auto-configuring exploration-exploitation tradeoff in evolutionary computation via deep reinforcement learning}. In \bibinfo{booktitle}{\emph{Proceedings of the Genetic and Evolutionary Computation Conference}}. \bibinfo{pages}{1497--1505}.
\newblock


\bibitem[Mnih(2013)]%
        {mnih2013playing}
\bibfield{author}{\bibinfo{person}{Volodymyr Mnih}.} \bibinfo{year}{2013}\natexlab{}.
\newblock \showarticletitle{Playing atari with deep reinforcement learning}.
\newblock \bibinfo{journal}{\emph{arXiv preprint arXiv:1312.5602}} (\bibinfo{year}{2013}).
\newblock


\bibitem[Nguyen et~al\mbox{.}(2025a)]%
        {nguyen2025importance}
\bibfield{author}{\bibinfo{person}{Tai Nguyen}, \bibinfo{person}{Phong Le}, \bibinfo{person}{Andr\'{e} Biedenkapp}, \bibinfo{person}{Carola Doerr}, {and} \bibinfo{person}{Nguyen Dang}.} \bibinfo{year}{2025}\natexlab{a}.
\newblock \showarticletitle{On the Importance of Reward Design in Reinforcement Learning-based Dynamic Algorithm Configuration: A Case Study on OneMax with $(1+(\lambda,\lambda))$-GA}. In \bibinfo{booktitle}{\emph{Proceedings of the Genetic and Evolutionary Computation Conference}} (NH Malaga Hotel, Malaga, Spain) \emph{(\bibinfo{series}{GECCO '25})}. \bibinfo{publisher}{Association for Computing Machinery}, \bibinfo{address}{New York, NY, USA}, \bibinfo{pages}{1162–1171}.
\newblock
\showISBNx{9798400714658}
\href{https://doi.org/10.1145/3712256.3726395}{doi:\nolinkurl{10.1145/3712256.3726395}}


\bibitem[Nguyen et~al\mbox{.}(2025b)]%
        {source}
\bibfield{author}{\bibinfo{person}{Tai Nguyen}, \bibinfo{person}{Phong Le}, \bibinfo{person}{Carola Doerr}, {and} \bibinfo{person}{Nguyen Dang}.} \bibinfo{year}{2025}\natexlab{b}.
\newblock \bibinfo{howpublished}{\url{https://github.com/taindp98/OneMax-MPDAC.git}}.
\newblock


\bibitem[Rasheed et~al\mbox{.}(2020)]%
        {rasheed2020deep}
\bibfield{author}{\bibinfo{person}{Faizan Rasheed}, \bibinfo{person}{Kok-Lim~Alvin Yau}, \bibinfo{person}{Rafidah~Md Noor}, \bibinfo{person}{Celimuge Wu}, {and} \bibinfo{person}{Yeh-Ching Low}.} \bibinfo{year}{2020}\natexlab{}.
\newblock \showarticletitle{Deep reinforcement learning for traffic signal control: A review}.
\newblock \bibinfo{journal}{\emph{IEEE Access}}  \bibinfo{volume}{8} (\bibinfo{year}{2020}), \bibinfo{pages}{208016--208044}.
\newblock


\bibitem[Rechenberg(1973)]%
        {Rechenberg}
\bibfield{author}{\bibinfo{person}{Ingo Rechenberg}.} \bibinfo{year}{1973}\natexlab{}.
\newblock \bibinfo{booktitle}{\emph{Evolutionsstrategie}}.
\newblock \bibinfo{publisher}{Friedrich Fromman Verlag (G{\"u}nther Holzboog {KG})}, \bibinfo{address}{Stuttgart}.
\newblock


\bibitem[Schulman et~al\mbox{.}(2017)]%
        {schulman2017proximal}
\bibfield{author}{\bibinfo{person}{John Schulman}, \bibinfo{person}{Filip Wolski}, \bibinfo{person}{Prafulla Dhariwal}, \bibinfo{person}{Alec Radford}, {and} \bibinfo{person}{Oleg Klimov}.} \bibinfo{year}{2017}\natexlab{}.
\newblock \showarticletitle{Proximal policy optimization algorithms}.
\newblock \bibinfo{journal}{\emph{arXiv preprint arXiv:1707.06347}} (\bibinfo{year}{2017}).
\newblock


\bibitem[Schumer and Steiglitz(1968)]%
        {SchumerS68}
\bibfield{author}{\bibinfo{person}{Michael~A. Schumer} {and} \bibinfo{person}{Kenneth Steiglitz}.} \bibinfo{year}{1968}\natexlab{}.
\newblock \showarticletitle{Adaptive step size random search}.
\newblock \bibinfo{journal}{\emph{IEEE Trans. Automat. Control}}  \bibinfo{volume}{13} (\bibinfo{year}{1968}), \bibinfo{pages}{270--276}.
\newblock


\bibitem[Sharma et~al\mbox{.}(2019)]%
        {sharma2019deep}
\bibfield{author}{\bibinfo{person}{Mudita Sharma}, \bibinfo{person}{Alexandros Komninos}, \bibinfo{person}{Manuel L{\'o}pez-Ib{\'a}{\~n}ez}, {and} \bibinfo{person}{Dimitar Kazakov}.} \bibinfo{year}{2019}\natexlab{}.
\newblock \showarticletitle{Deep reinforcement learning based parameter control in differential evolution}. In \bibinfo{booktitle}{\emph{Proceedings of the genetic and evolutionary computation conference}}. \bibinfo{pages}{709--717}.
\newblock


\bibitem[Speck et~al\mbox{.}(2021)]%
        {speck2021learning}
\bibfield{author}{\bibinfo{person}{David Speck}, \bibinfo{person}{Andr{\'e} Biedenkapp}, \bibinfo{person}{Frank Hutter}, \bibinfo{person}{Robert Mattm{\"u}ller}, {and} \bibinfo{person}{Marius Lindauer}.} \bibinfo{year}{2021}\natexlab{}.
\newblock \showarticletitle{Learning heuristic selection with dynamic algorithm configuration}. In \bibinfo{booktitle}{\emph{Proceedings of the International Conference on Automated Planning and Scheduling}}, Vol.~\bibinfo{volume}{31}. \bibinfo{pages}{597--605}.
\newblock


\bibitem[Sutton and Barto(1998)]%
        {712192}
\bibfield{author}{\bibinfo{person}{R.S. Sutton} {and} \bibinfo{person}{A.G. Barto}.} \bibinfo{year}{1998}\natexlab{}.
\newblock \showarticletitle{Reinforcement Learning: An Introduction}.
\newblock \bibinfo{journal}{\emph{IEEE Transactions on Neural Networks}} \bibinfo{volume}{9}, \bibinfo{number}{5} (\bibinfo{year}{1998}), \bibinfo{pages}{1054--1054}.
\newblock
\href{https://doi.org/10.1109/TNN.1998.712192}{doi:\nolinkurl{10.1109/TNN.1998.712192}}


\bibitem[Tavakoli et~al\mbox{.}(2018)]%
        {tavakoli2018action}
\bibfield{author}{\bibinfo{person}{Arash Tavakoli}, \bibinfo{person}{Fabio Pardo}, {and} \bibinfo{person}{Petar Kormushev}.} \bibinfo{year}{2018}\natexlab{}.
\newblock \showarticletitle{Action branching architectures for deep reinforcement learning}. In \bibinfo{booktitle}{\emph{Proceedings of the aaai conference on artificial intelligence}}, Vol.~\bibinfo{volume}{32}.
\newblock


\bibitem[Tessari and Iacca(2022)]%
        {tessari2022reinforcement}
\bibfield{author}{\bibinfo{person}{Michele Tessari} {and} \bibinfo{person}{Giovanni Iacca}.} \bibinfo{year}{2022}\natexlab{}.
\newblock \showarticletitle{Reinforcement learning based adaptive metaheuristics}. In \bibinfo{booktitle}{\emph{Proceedings of the Genetic and Evolutionary Computation Conference Companion}}. \bibinfo{pages}{1854--1861}.
\newblock


\bibitem[Van~Hasselt et~al\mbox{.}(2016)]%
        {van2016deep}
\bibfield{author}{\bibinfo{person}{Hado Van~Hasselt}, \bibinfo{person}{Arthur Guez}, {and} \bibinfo{person}{David Silver}.} \bibinfo{year}{2016}\natexlab{}.
\newblock \showarticletitle{Deep reinforcement learning with double q-learning}. In \bibinfo{booktitle}{\emph{Proceedings of the AAAI conference on artificial intelligence}}, Vol.~\bibinfo{volume}{30}.
\newblock


\bibitem[Vermetten et~al\mbox{.}(2024)]%
        {VermettenLRBD24}
\bibfield{author}{\bibinfo{person}{Diederick Vermetten}, \bibinfo{person}{Johannes Lengler}, \bibinfo{person}{Dimitri Rusin}, \bibinfo{person}{Thomas B{\"{a}}ck}, {and} \bibinfo{person}{Carola Doerr}.} \bibinfo{year}{2024}\natexlab{}.
\newblock \showarticletitle{Empirical Analysis of the Dynamic Binary Value Problem with IOHprofiler}. In \bibinfo{booktitle}{\emph{Proc. of Parallel Problem Solving from Nature (PPSN)}} \emph{(\bibinfo{series}{LNCS}, Vol.~\bibinfo{volume}{15149})}. \bibinfo{publisher}{Springer}, \bibinfo{pages}{20--35}.
\newblock
\href{https://doi.org/10.1007/978-3-031-70068-2\_2}{doi:\nolinkurl{10.1007/978-3-031-70068-2\_2}}


\bibitem[Watkins and Dayan(1992)]%
        {watkins1992q}
\bibfield{author}{\bibinfo{person}{Christopher~JCH Watkins} {and} \bibinfo{person}{Peter Dayan}.} \bibinfo{year}{1992}\natexlab{}.
\newblock \showarticletitle{Q-learning}.
\newblock \bibinfo{journal}{\emph{Machine learning}}  \bibinfo{volume}{8} (\bibinfo{year}{1992}), \bibinfo{pages}{279--292}.
\newblock


\bibitem[Xue et~al\mbox{.}(2022)]%
        {xue2022multi}
\bibfield{author}{\bibinfo{person}{Ke Xue}, \bibinfo{person}{Jiacheng Xu}, \bibinfo{person}{Lei Yuan}, \bibinfo{person}{Miqing Li}, \bibinfo{person}{Chao Qian}, \bibinfo{person}{Zongzhang Zhang}, {and} \bibinfo{person}{Yang Yu}.} \bibinfo{year}{2022}\natexlab{}.
\newblock \showarticletitle{Multi-agent dynamic algorithm configuration}.
\newblock \bibinfo{journal}{\emph{Advances in Neural Information Processing Systems}}  \bibinfo{volume}{35} (\bibinfo{year}{2022}), \bibinfo{pages}{20147--20161}.
\newblock


\bibitem[Yi and Qu(2023)]%
        {yi2023automated}
\bibfield{author}{\bibinfo{person}{Wenjie Yi} {and} \bibinfo{person}{Rong Qu}.} \bibinfo{year}{2023}\natexlab{}.
\newblock \showarticletitle{Automated design of search algorithms based on reinforcement learning}.
\newblock \bibinfo{journal}{\emph{Information Sciences}}  \bibinfo{volume}{649} (\bibinfo{year}{2023}), \bibinfo{pages}{119639}.
\newblock


\bibitem[Yi et~al\mbox{.}(2022)]%
        {yi2022automated}
\bibfield{author}{\bibinfo{person}{Wenjie Yi}, \bibinfo{person}{Rong Qu}, \bibinfo{person}{Licheng Jiao}, {and} \bibinfo{person}{Ben Niu}.} \bibinfo{year}{2022}\natexlab{}.
\newblock \showarticletitle{Automated design of metaheuristics using reinforcement learning within a novel general search framework}.
\newblock \bibinfo{journal}{\emph{IEEE Transactions on Evolutionary Computation}} \bibinfo{volume}{27}, \bibinfo{number}{4} (\bibinfo{year}{2022}), \bibinfo{pages}{1072--1084}.
\newblock


\bibitem[Zahavy et~al\mbox{.}(2018)]%
        {zahavy2018learn}
\bibfield{author}{\bibinfo{person}{Tom Zahavy}, \bibinfo{person}{Matan Haroush}, \bibinfo{person}{Nadav Merlis}, \bibinfo{person}{Daniel~J Mankowitz}, {and} \bibinfo{person}{Shie Mannor}.} \bibinfo{year}{2018}\natexlab{}.
\newblock \showarticletitle{Learn what not to learn: Action elimination with deep reinforcement learning}.
\newblock \bibinfo{journal}{\emph{Advances in neural information processing systems}}  \bibinfo{volume}{31} (\bibinfo{year}{2018}).
\newblock


\end{thebibliography}

\end{document}